\pdfoutput=1

\documentclass[11pt]{article}

\usepackage{ACL2023}

\usepackage{times}
\usepackage{latexsym}
\usepackage{booktabs} 
\usepackage{CJKutf8}
\usepackage{tabularx}
\usepackage{subcaption} 
\usepackage[T1]{fontenc}

\usepackage[utf8]{inputenc}

\usepackage{microtype}

\usepackage{inconsolata}

\usepackage{graphicx}

\title{RoleCraft-GLM: Advancing Personalized Role-Playing in Large Language Models}

\author{
  Meiling Tao$^{1}$\thanks{$^{*}$Equal contribution.}, 
  Xuechen Liang$^{2{*}}$, 
  Tianyu Shi$^{3}$\thanks{$^{*}$Corresponding author: \texttt{ty.shi@mail.utoronto.ca}}, 
  Lei Yu$^{3†}$,
  Yiting Xie$^{4}$ \\
  \begin{tabular}[t]{c}
    $^1$Guangdong University of Technology, Guangzhou, China \\
    $^2$East China Jiaotong University, Nanchang, China\\
    $^3$ University of Toronto, Toronto, Canada \\
    $^4$Genfun.ai, Beijing, China \\
  \end{tabular}
}

\begin{document}
\maketitle
\begin{abstract}

The development of large language models(LLMs) has initiated a new chapter in complex tasks such as role-playing, enhancing user interaction experiences by enabling models to imitate various characters.However, LLMs are somewhat lacking in their ability to portray lesser-known characters, especially in aspects of dialogue delivery and scriptwriting skills. To this end, we aim to swiftly acquire essential language skills for character development, greatly enhancing role-playing comfort. In this work, we present RoleCraft, an innovative framework designed to enrich personalized role-playing experiences. Central to this framework is RoleInstruct, a distinctive dataset featuring emotional annotations, transitioning from traditional celebrity-focused roles to more authentic, daily non-celebrity roles,each accompanied by carefully crafted character descriptions. We combined RoleInstruct with open-source instructions from the general domain, employing a hybrid instruction tuning strategy to create RoleCraft-GLM. Experiments in role-playing demonstrate that our model excels in generating dialogue that accurately reflects character traits and emotions, outperforming most mainstream LLMs, including GPT-4\footnote{Access models, demos at https://github.com/tml2002/RoleCraft}.

\end{abstract}

\section{Introduction}

\begin{figure*}[htbp]
    \centering
    \includegraphics[width=\linewidth]{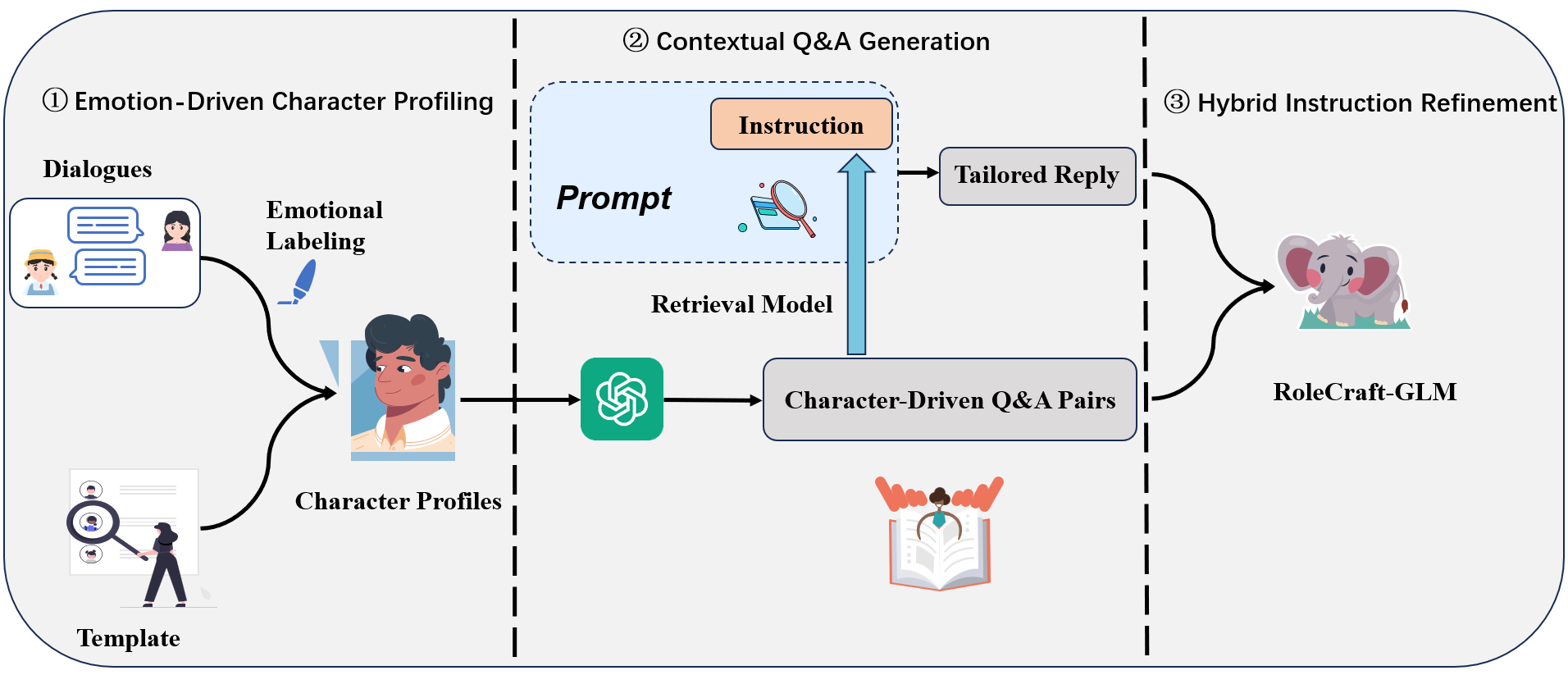} 
    \caption{Overview of the RoleCraft-GLM framework: (1) Dialog datasets annotated with emotions are utilized to construct role profiles embodying distinct emotional features. (2) The generation of Q\&A pairs, based on context and known character traits, ensures that dialogues are consistent with the character profiles. (3) A hybrid approach of generic and character-specific instructions is used to train the GLM for various dialog scenarios.}
    \label{fig:example}
\end{figure*}

Large Language Models (LLMs) have emerged as pivotal in understanding and generating natural language, often surpassing human capabilities in some language reasoning tasks. However, existing open-source LLMs, primarily trained in general domains, lack the specialized optimization needed for nuanced role-playing tasks, indicating a need for further customization to effectively meet specific role-playing requirements. Furthermore, advanced LLMs like GPT-4~\cite{openai:gpt4} demonstrate improved role-playing abilities due to extensive training and sophisticated algorithms, but as a closed-source model, it poses practical challenges. These challenges include higher costs for API use, limited fine-tuning opportunities for specific role-playing contexts, and context window size restrictions that may affect the continuity and depth of dialogues in complex role-playing scenarios.In light of this, a natural question arises: \textit{Can we meticulously train and tailor open-source LLMs to achieve role-playing capabilities comparable to closed-source LLMs, while also enhancing their ability to meet individual user needs?}

Previous efforts in AI role-playing have primarily centered on celebrity figures~\cite{shao2023character,li2023chatharuhi}, heavily reliant on predefined domain knowledge and often lacking in relevance to daily life and personalization. While there have been improvements in learning from past interactions and adapting to user needs, these advancements have not sufficiently addressed the challenges of achieving truly personalized AI interactions~\cite{chen2023palr}. Existing role-playing models still lack the necessary depth and individual specificity to accurately reflect the diverse and nuanced experiences of everyday users.

To enhance the role-playing capabilities of LLMs, there are two main challenges: (1) Limited Personalization in Generic AI Models:  Many AI systems currently rely on generic models that frequently fail to address the specific and nuanced needs of diverse user groups. This limitation undermines the overall user experience and restricts the practical application of AI in specialized scenarios~\cite{Ackerman2022High-quality}. (2) Gap in Sophisticated Role-Playing Capabilities: In the realm of conversational agents, despite significant achievements in providing high-level responses across a variety of dialogues and enhancing human-AI interactions~\cite{bender2020climbing}, there remains a noticeable gap in AI's capacity for sophisticated role-playing. Current models often lack the depth and adaptability required for truly personalized and nuanced role-play experiences. 

In this paper, as illustrated in the figure~\ref{fig:example}, we introduce the RoleCraft framework designed to enhance personalized role-playing experiences with LLMs. Moving beyond traditional celebrity-focused characters, we focus on diverse, non-celebrity personas, each with unique emotional annotations. This approach aims to enrich realism and emotional depth in language interactions. We compiled a novel dataset encompassing a wide array of real-world dialogues, with careful consideration for personal privacy and copyright laws. Our data analysis highlights the potential benefits of integrating emotional labels in dialogue datasets for improved natural language processing. We conducted comparative experiments using models like ChatGLM3, fine-tuned with the Low-Rank Adaptation (LoRA) method, to assess RoleCraft-GLM's effectiveness in producing nuanced and character-consistent dialogues.

The main contributions of our work are as follows:
\begin{itemize}
    \item We introduce a novel RoleInstruct dataset, which centers on non-celebrity characters, each characterized by unique emotional annotations. 
    \item We develop RoleCraft, a novel framework that integrates a more detailed approach to personal role data in training strategies. RoleCraft includes a fine-grained method for character portrayal, emphasizing emotional depth, and fostering contextual awareness in dialogue generation.
    \item RolePlay-GLM, our fine-tuned model, demonstrates promising performance against current state-of-the-art models, excelling in dialogue authenticity, emotional accuracy, and contextual relevance as per comprehensive evaluations.
\end{itemize}




\section{Related Work}

\subsection{Role-Playing}

The evolution of role-playing in AI~\cite{wu2023tidybot}, marked by the transition from basic text-based interactions to intricate character simulations~\cite{park2023generative}, reflects the strides made in natural language processing and AI technologies~\cite{mysore2023pearl}. Initially, AI role-playing systems offered only fundamental exchanges, limited in their ability to craft dialogues with emotional depth and contextual relevance. With the emergence of advanced models such as GPT-3~\cite{brown2020language}, LLaMA~\cite{touvron2023llama}, and ChatGLM, there was a notable enhancement in AI's capability for engaging in more sophisticated, context-aware conversations. Yet, these improvements also underscored a significant gap in personalization for role-playing applications. Predominantly, LLMs trained on wide-ranging, generic datasets fell short in handling scenarios that demanded a deeper understanding of nuanced emotional nuances~\cite{radford2018improving} and specific character traits. To address these shortcomings, we meticulously develop the RoleCraft framework. It stands out with its unique dataset, focusing on diverse, non-celebrity personas enriched with detailed emotional annotations~\cite{bender2020climbing}. This dataset is key to overcoming previous limitations, facilitating a new level of personalization and emotional intricacy in AI role-playing interactions.

\subsection{Personalization of LLMs}

The recent strides in LLMs, particularly in understanding user context and preferences~\cite{wang2023learning,abbasian2023conversational}, have significantly propelled the personalization aspect of AI interactions~\cite{lee2022evaluating,subhash2023can}. Previous works~\cite{shanahan2023role,li2023camel,chen2023autoagents} have demonstrated the potential of LLMs in mimicking specific fictional characters and simulating complex human behaviors. However, these models often face challenges in achieving a deep level of personalization and emotional richness that aligns precisely with individual user contexts and needs~\cite{milkowski2022multitask}. Additionally, while these models are adept at simulating personalities or historical figures, they may not effectively handle the subtleties of user-specific emotional responses or cater to nuanced personal preferences. Our work aims to address these gaps by building upon these advancements and offering a framework that focuses on a more granular level of personalization. We propose novel methods for emotional and context-specific interaction, ensuring that our model can adapt and respond more accurately to individual user scenarios.

\begin{figure*}[htbp]
    \centering
    \vspace{-20pt}
    \includegraphics[width=0.8\linewidth]{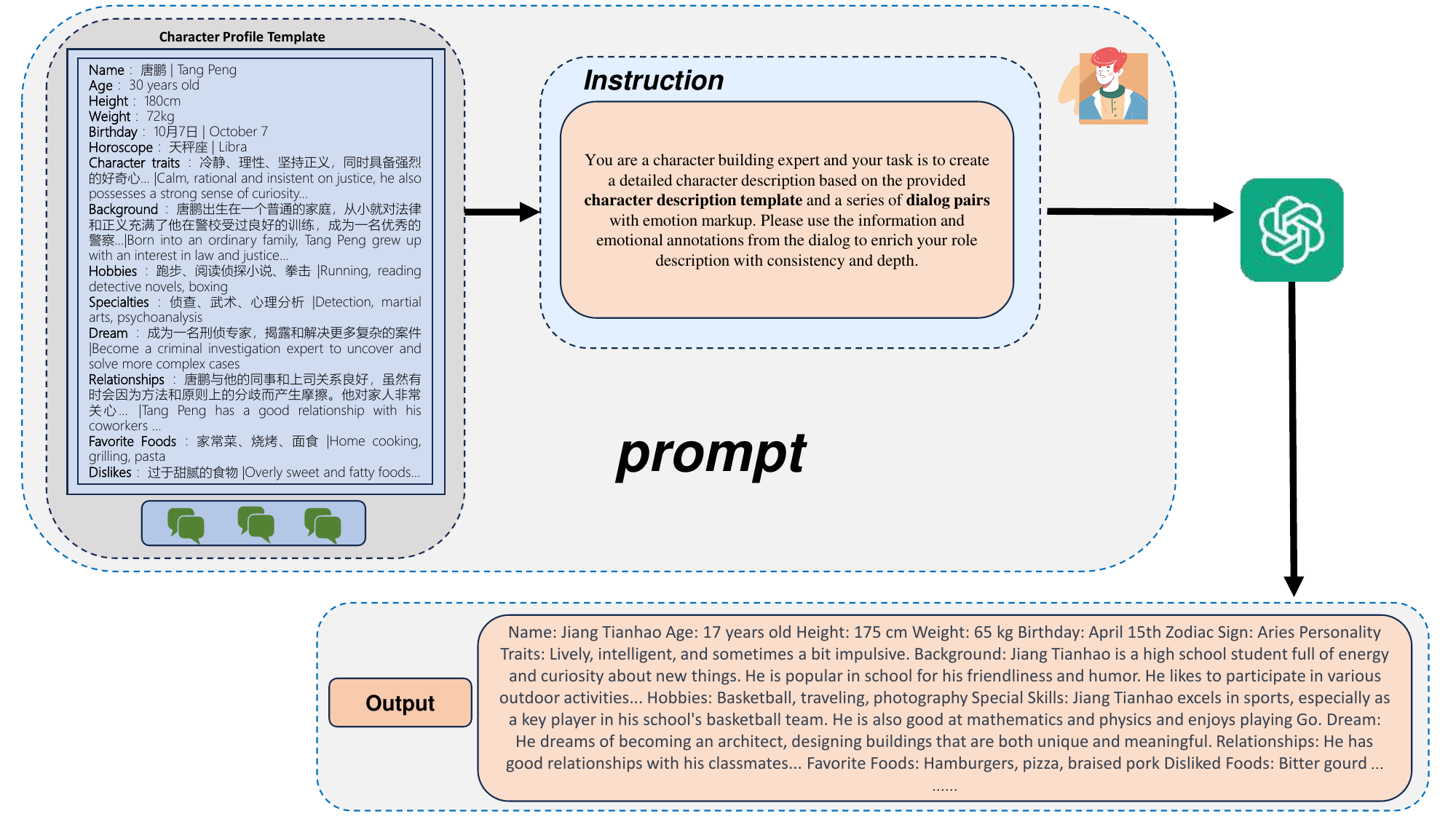}
    \caption{An example of crafting a detailed character portrayal. By leveraging a character description template and a dialogue dataset with emotional annotations.(The instruction and output have been translated into English)}
    \label{fig:example1}
\end{figure*}

\section{Methodology}


As shown in Figure~\ref{fig:example}, the RoleCraft framework, rooted in `Role' and `Craft', represents our approach to enhancing AI role-playing. `Role' emphasizes creating distinct, multi-dimensional characters, each with unique personality traits and emotional depths. `Craft' involves the intricate process of constructing dialogues that genuinely reflect these character traits, thereby adding depth and realism to conversations.See Appendix~\ref{sec:appendixA}  for more details.

\subsection{Overall Framework}
Our methodology uniquely advances the capabilities of LLMs in role-playing. Setting ourselves apart from approaches such as RoleLLM~\cite{wang2023rolellm}, we focus on an innovative integration of fine-grained character portrayal, profound emotional depth, and heightened contextual awareness in dialogue generation. This approach differentiates our work from existing models and addresses challenges in a novel way, enhancing how LLMs can be utilized for creating more realistic and engaging role-playing scenario.

\subsubsection*{Emotion-Driven Character Profiling}

To address the challenges of limited emotional diversity and unconvincing character portrayals in dialogues generated by LLMs, we adopt a detailed emotion classification strategy. This approach involves meticulously annotating emotions within the dialogue dataset, thereby steering the GPT-4 to craft character profiles that mirror these identified emotions. Consider a character who displays a spectrum of emotions from joy to disappointment. Marking these diverse emotional states allows for a natural and fluid transition in their dialogues within a single scene, effectively capturing the complexity and dynamism of human emotions. Our approach challenges the LLMs to accurately depict these emotional shifts, ensuring that the dialogues genuinely represent the intricate and ever-evolving nature of human emotions, thus enhancing the overall user interaction experience.

\subsubsection*{Contextual Q\&A Generation} 

To address the challenge of context-irrelevant responses~\cite{feng2023learning,ye2022reflecting}, which is a common issue in dialogue systems~\cite{ni2023recent} where interactions often lack relevance to the ongoing scenario or character specifics~\cite{mitsuda2022investigating}, we employ GPT-4 to generate contextually coherent Q\&A pairs. For example, when a character faces a dilemma, the system is designed to produce queries and responses that align with the character's established traits, such as indecisiveness and anxiety, thereby maintaining the authenticity of the dialogue in relation to the character's profile.

\subsubsection*{Hybrid Instruction-Based GLM Refinement }

Our methodology employs a hybrid training approach that seamlessly integrates general instructions with character-specific Q\&A pairs. This strategy is carefully crafted to strike a balance between the flexibility required for dynamic dialogue generation and the need to uphold character integrity. In practical terms, this means equipping the LLMs to adeptly navigate a spectrum of conversational scenarios. These range from broad, general interactions to more intricate exchanges that demand responses finely tuned to the unique profiles of individual characters. By training the LLM with this diverse mix of inputs, the model becomes proficient in handling various situational dialogues, accurately reflecting each character's distinct attributes and the specific subtleties of the conversation. As a result, this hybrid training method fosters the creation of dialogues that are both contextually adaptive and consistent with the characters' distinct personalities.


\subsection{Semantic-Enhanced Retrieval Optimization}
Addressing the issue of inaccurate and semantically irrelevant information retrieval in dialogues, we adopt the BGE\footnote{https://github.com/FlagOpen/FlagEmbedding} retrieval method. BGE is an efficient Chinese and English semantic vector model that ensures the accuracy of responses, especially when dealing with sensitive topics, and remains semantically sensitive to the context, significantly enhancing the quality of interaction~\cite{xiao2023c}. This familiarity allows models to generate dialogue based on a wealth of pre-existing knowledge. In contrast, modern datasets prioritize the nuanced portrayal of personal and everyday characters. These datasets are derived from diverse sources, including real chat logs, customer service interactions, and fictional narratives from less mainstream media. Such characters might include a typical office worker dealing with daily stressors or a mother showing love and responsibility in a family setting. The dialogues here involve specific, real-life scenarios, such as office interactions or typical family conversations, which lack the broad pre-existing knowledge base associated with public figures.

\subsection{Compared to Existing Practice}

Our methodology diverges from conventional prompt engineering~\cite{white2023prompt} and fine-tuning~\cite{howard2018universal} We specialize in creating diverse, intricate character interactions that enhance role-playing scenarios' depth and authenticity. Our unique approach combines emotion-driven character profiles with contextually coherent Q\&A generation, fostering realistic and engaging interactions. Additionally, the integration of semantic-enhanced retrieval methods ensures both accuracy and relevance, setting our approach apart in its complexity and user engagement potential.

\begin{figure*}
\centering
    \vspace{-20pt}
\begin{minipage}[b]{0.48\linewidth}
\centering
\includegraphics[width=0.8\linewidth]{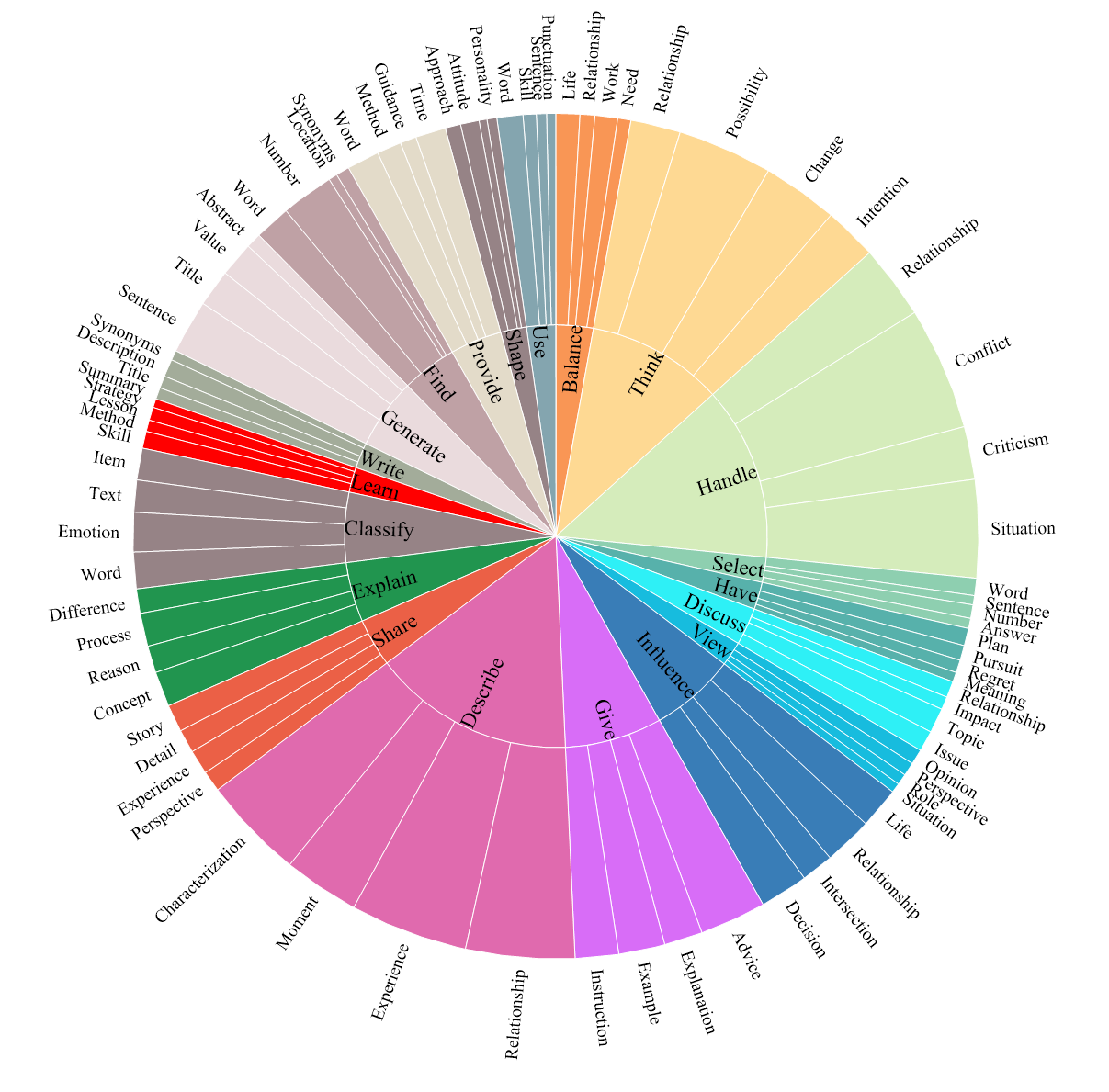}
\caption{Verb-noun structure of Instructions. The inner circle representing the top 20 verbs and the outer circle listing the direct noun objects.}
\label{fig:image}
\end{minipage}
\hfill
\begin{minipage}[b]{0.48\linewidth}
\begin{minipage}[b]{\linewidth}
\centering
\includegraphics[width=0.65\linewidth]{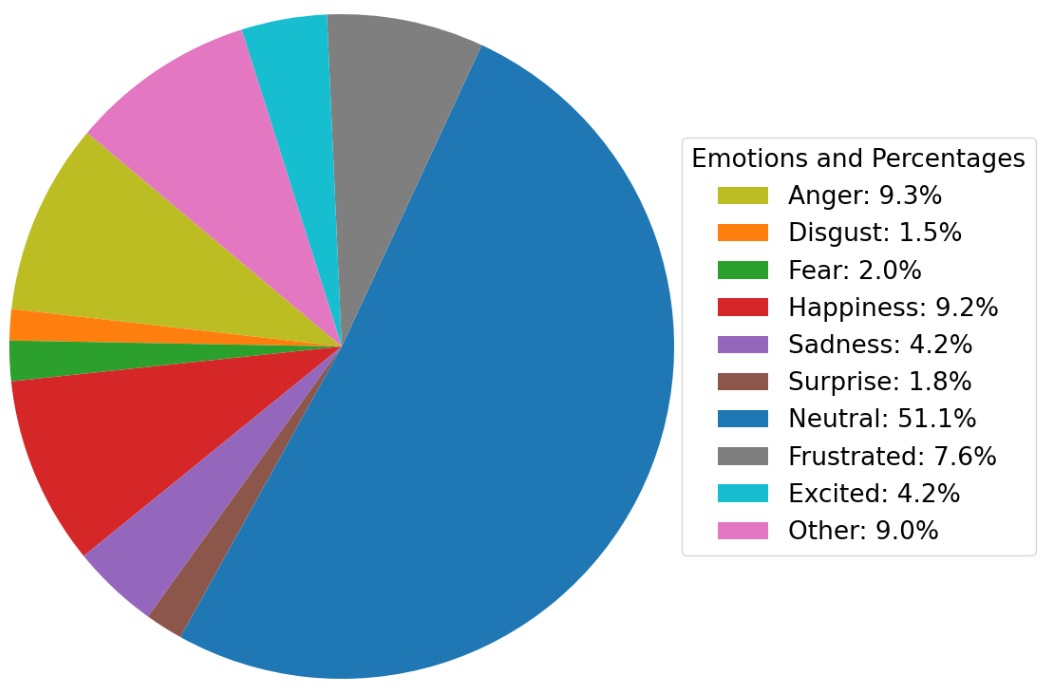}
    \caption{ Emotion distributions in dialogues}
    \label{Emotion distributions in dialogues}
\end{minipage}\\[10pt] 
\begin{minipage}[b]{\linewidth}
\centering
\includegraphics[width=0.7\linewidth]{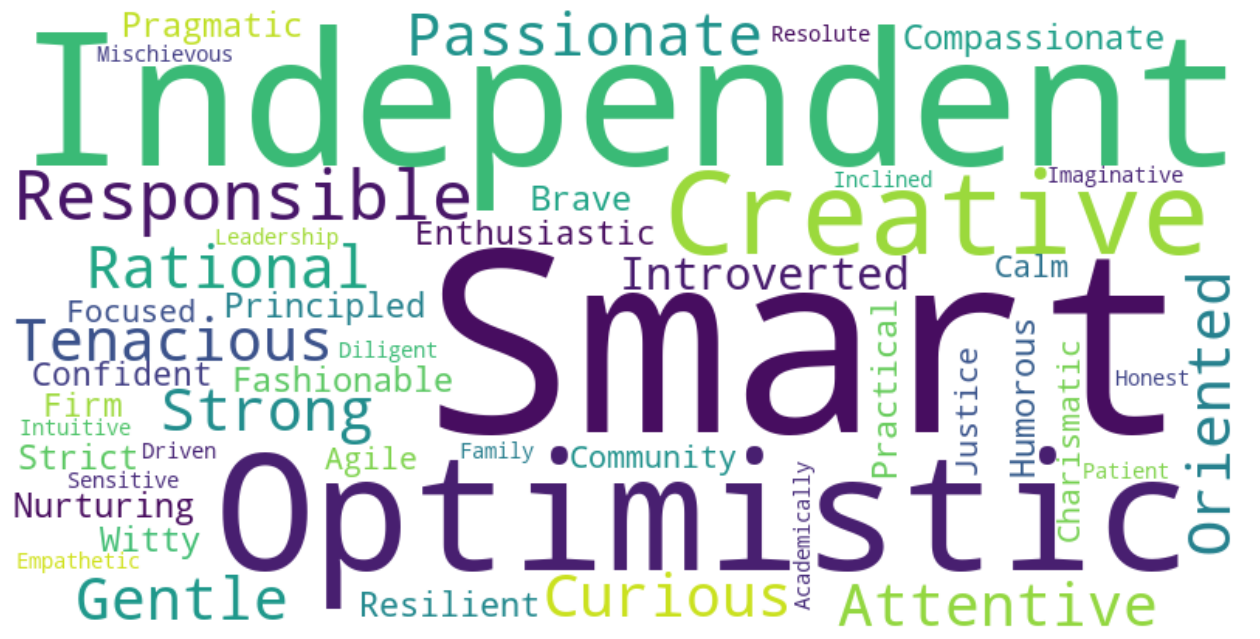}
\caption{Word Cloud Of Character Personality Traits}
\label{Word Cloud Of Character Personality Traits}
\end{minipage}
\end{minipage}
\end{figure*} 

\begin{table}[h]
        \centering
            \captionof{table}{Statistics of datasets}
          \begin{tabular}{lc}
          \hline
           \textbf{Category} & \textbf{Value}\\
        \hline
            \# Total Dialogues & 48,677\\
               Avg.round of dialogues & 14.85\\
        \hline
            \# Characters & 28\\
               Character Personality Traits & 45 \\
               Avg.length of profile & 382.15\\
        \hline
           \# Instructions & 43,358\\
              Character-specific instructions & 13,778\\
              General instructions & 29,580\\
              Avg. instruction length & 27.68\\
        \hline
           \# Response & 161,678\\
              Character-specific response & 13,778\\
              General response & 147,900\\
              Avg.response length & 33.29\\
              \hline
        \end{tabular}

        \label{tab:table}
\end{table}   

\section{Experiments}
Building on previous work~\cite{tao2023rolecraft}, we introduced new evaluation methods and made comprehensive enhancements to the experimental process, conducting an in-depth analysis of model performance changes in specific scenarios. We anticipate that fine-tuning our model using a specifically designed dataset for role-playing will result in superior performance in character portrayal compared to baseline models. Through this specialized training, we expect our model to accurately capture and express the intricacies of character-specific language styles and emotional responses, surpassing baseline models that may lack such targeted training. Our experiments aim to validate this hypothesis and showcase the advanced capabilities of our model in role-playing tasks.

\subsection{Dataset}

In the evaluation of LLMs~\cite{chen2021evaluating,chang2023survey}, the role of datasets is paramount, particularly in language processing and character portrayal. Traditional role playing datasets predominantly highlight eminent figures, such as the legendary Sun Wukong, whose familiar attributes and stories are widely acknowledged, facilitating model development~\cite{Sabadoš2021AUTOMATSKO}. However, these datasets often neglect the finer details and emotional complexity of lesser-known or everyday characters, leading to a representation gap~\cite{Rolf2021Representation}. Our unique dataset bridges this gap by focusing on the rich, nuanced depiction of ordinary individuals. It involves an in-depth exploration and portrayal of each character's distinct personality traits and emotional depths, delving into aspects usually overshadowed in dominant narratives. 

In constructing our dataset, we design 28 unique and personalized Chinese characters to mirror a wide spectrum of real-world dialogues. These characters ranged from everyday individuals to specialized roles such as intelligent customer service agents and research assistants.  Table~\ref{tab:table} provides basic statistics for RoleInstruct. Our diverse data sources included social media interactions, film and television scripts, and customer service dialogues. We emphasized personal privacy and copyright law compliance, ensuring all data was cleansed and anonymized.

 Figures~\ref{fig:image} and~\ref{Word Cloud Of Character Personality Traits} visually demonstrate the diversity of RoleInstruct, both in terms of linguistic usage and emotional expression. The verb-noun diagram provides insights into the varied narrative contexts, while the word cloud delves into the breadth of emotional and personality traits present.

In addition, we annotate each dataset entry with emotion labels to capture characters' distinct emotional traits, adding an emotional layer to model training. We use Ekman's "Six Basic Emotions Theory"~\cite{ekman1992argument} to label utterances and included additional emotions like neutral, excited, and depressed, totaling ten categories. The use of emotion labels in dialogue datasets has been proven to enhance natural language processing by improving response retrieval and emotional relevance ~\cite{Zhou2017Single}. These labels also enrich conversational analysis and aid in building natural dialogue systems ~\cite{Bothe2019Enriching}. See Appendix~\ref{sec:appendixB}  for more details.

\subsection{Implementation Settings}

We assess the ChatGLM3 model, enhancing its performance on specific datasets using the Low-Rank Adaptation (LoRA) fine-tuning method~\cite{hu2021lora}. LoRA's precision in fine-tuning, essential for handling personalized and emotionally rich content, maintains the model's core capabilities while adapting to new data features. Please refer to Appendix~\ref{sec:appendixC} for more details.

\subsection{Baselines}

We benchmark our RoleCraft-GLM's performance against industry standards such as GPT-3.5 and GPT-4, and leading Chinese dialogue generation technologies like ChatGLM2\footnote{https://github.com/THUDM/ChatGLM2-6B} and ChatGLM3, along with Baichuan2~\cite{yang2023baichuan}, Qwen~\cite{qwen}, InternLM~\cite{team2023internlm}, SparkDesk\footnote{https://xinghuo.xfyun.cn/} and Xverse\footnote{https://github.com/xverse-ai/XVERSE-13B}. Additionally, we introduce another role-playing baseline, RoleLLM, which utilizes LoRA for fine-tuning on a specific dataset.

\subsection{Evaluation criteria}

\subsubsection{Rouge-L Score}

    A commonly used metric~\cite{lin2004rouge} for evaluating natural language generation, measuring the overlap between model-generated text and real (ground truth) text. We focused on average score (Avg), general instruction response (RAW), role-playing speaking style (CUS), and specific role knowledge (SPE).
    
\subsubsection{GPT Score}
    We use the GPT-4 score~\cite{fu2023gptscore} to evaluate the average rank of models on different dialogue-generating tasks.Please refer to Appendix~\ref{sec:appendixD} for more details.
  
\subsubsection{RPCS}
    We introduce Role-Playing Cosine Similarity (RPCS) as a new evaluation standard to accurately assess model performance in role-playing scenarios. RPCS evaluates the emotional and content consistency between the model-generated responses and the expected responses by calculating the cosine similarity between two text segments. Concurrently, we use OpenAI's Text-Embedding-Ada002 model to convert texts into vector representations, enabling deeper capture of semantic features of texts, thereby enhancing the precision of our assessment. 

\subsubsection{Human Evaluation}
To effectively assess role-playing agents, we engaged 12 annotators who rated model-generated responses using three key metrics on a five-point scale: Communication Effectiveness, Consistency, and Expressive Diversity~\cite{zhang2021dynaeval,mesgar2020improving,tu2024charactereval}.
\begin{itemize}
 \item \textbf{Communication Effectiveness(CE):} 
 Merging fluency and coherence, the metric evaluates the dialogue's smoothness, grammatical accuracy, logical consistency, and contextual relevance.  
 \item \textbf{Consistency:} 
 Encompassing knowledge and persona consistency, this metric assesses the accuracy of the agent's information relative to the role's background and the adherence to the character's traits and behaviors.  
 \item \textbf{Expressive Diversity(ED):} 
 Focused on evaluating the agent's creativity in dialogue generation.It's vital for assessing the agent's ability to enhance user experience and interaction engagement.
\end{itemize}

 \begin{table}[ht]
  \caption{Rouge-L Evaluation}
  \label{tab:freq}
  \begin{tabular}{lcccc}
  \hline
\textbf{Model} & \textbf{Avg} & \textbf{RAW}&\textbf{CUS}&\textbf{SPE}\\
   \hline
   GPT-3.5 & 0.4532 & 0.5569 & 0.5197 & 0.2831 \\
   GPT-4  & 0.4633 & \textbf{0.5661} & \textbf{0.5264} & 0.2973 \\
   ChatGLM2 & 0.4054 & 0.5104 & 0.4063 & 0.2996 \\
   ChatGLM3 & 0.4161 & 0.5218 & 0.4159 & 0.3108 \\
   Baichuan2 &0.4273 & 0.5308 & 0.4576 & 0.2934 \\
   Qwen &0.4264 & 0.5297 & 0.4617 & 0.2879 \\
   InternLM & 0.3947 & 0.4937 & 0.4076 & 0.2829 \\
   SparkDesk & 0.4288 & 0.5341 & 0.4510 & 0.3014\\
   Xverse & 0.4271 & 0.5180 & 0.4653 & 0.2981 \\
   RoleGLM & 0.4570 & 0.5255 & 0.5049 & 0.3406 \\
   \hline
   Ours & \textbf{0.4704} &0.5385 & 0.5154& \textbf{0.3573}\\
   \hline
\end{tabular}
\end{table}    

\begin{table*}[tbp]
\centering

\begin{minipage}[t]{0.48\linewidth} 
\centering
\caption{GPT-4 Evaluation}
\label{table3}
  \begin{tabular}{lc}
  \hline
    \textbf{Model} & \textbf{Avg. Ranking} \\
    \hline
    ChatGLM3 & 2.96 \\
    Baichuan2 & 4.57 \\
    Qwen & 5.71 \\
    InternLM &  6.29\\
    SparkDesk & 4.29 \\
    Xverse & 3.43 \\
    RoleGLM & 2.21\\
    \hline
    Ours & \textbf{1.43} \\
    \hline
  \end{tabular}
\end{minipage}
\hfill
\begin{minipage}[t]{0.48\linewidth} 
\centering
\caption{Results of point-wise evaluation}
\label{table4}
  \begin{tabular}{lcccc}
  \hline
\textbf{Model} & \textbf{Avg} & \textbf{CE}&\textbf{Consistency}&\textbf{ED}\\
   \hline
   GPT-3.5 & 2.95 & 3.08 & 3.88 & 1.89 \\
   GPT-4 & 3.33 & 3.50 & \textbf{4.26} & 2.23 \\
   ChatGLM2 & 2.56 & 3.12 & 2.65 & 1.91 \\
   ChatGLM3 & 2.92 & 3.36 & 3.32  & 2.07 \\
   Baichuan2 & 3.29 & 3.61 & 4.02 & 2.25 \\
   Qwen  & 3.17 & 3.27 & 3.97 & 2.27 \\
   InternLM & 2.50 & 2.93 & 2.40 & 2.16 \\
   SparkDesk & 3.03 & 3.52 & 3.44 & 2.13\\
   Xverse & 3.02 & 3.48 & 3.53 & 2.05 \\
   \hline
   Ours & \textbf{3.44 } &\textbf{ 3.75}& 4.14 & \textbf{2.43} \\
   \hline
\end{tabular}
\end{minipage}
\end{table*}

\begin{table*}[tbp]
\centering
\begin{minipage}[t]{0.48\linewidth}
\centering
  \caption{\centering{Comparing Emotion-Annotated and Non-Annotated Data}} 
\begin{tabular}{lc}
\hline
\textbf{Method} & \textbf{SPE}\\
\hline
RoleCraft-GLM(w/o emo) & 0.3362 \\
RoleCraft-GLM(w emo) & \textbf{0.3573} \\
\hline 
\end{tabular}
     \label{tab:freq3}    
\end{minipage}
\hfill
\begin{minipage}[t]{0.48\linewidth}
\centering
  \caption{\centering{Ablation study on the effect of different instructions}} 
\begin{tabular}{lcccc}
\hline
\textbf{RoleCraft-GLM} & \textbf{Avg} & \textbf{RAW}&\textbf{CUS}&\textbf{SPE} \\
\hline
- general  & 0.4311 & \textbf{0.5402} & \textbf{0.5219} & 0.2311 \\
- specific  & 0.4045 & 0.4387 & 0.4143 & \textbf{0.3606} \\
- hybrid & \textbf{0.4704} & 0.5385 & 0.5154& 0.3573 \\
\hline  
\end{tabular}
     \label{tab:freq4} 
\end{minipage}
\end{table*}

\subsection{Performance Analysis}

\begin{figure}
    \centering
    \includegraphics[width=\linewidth]{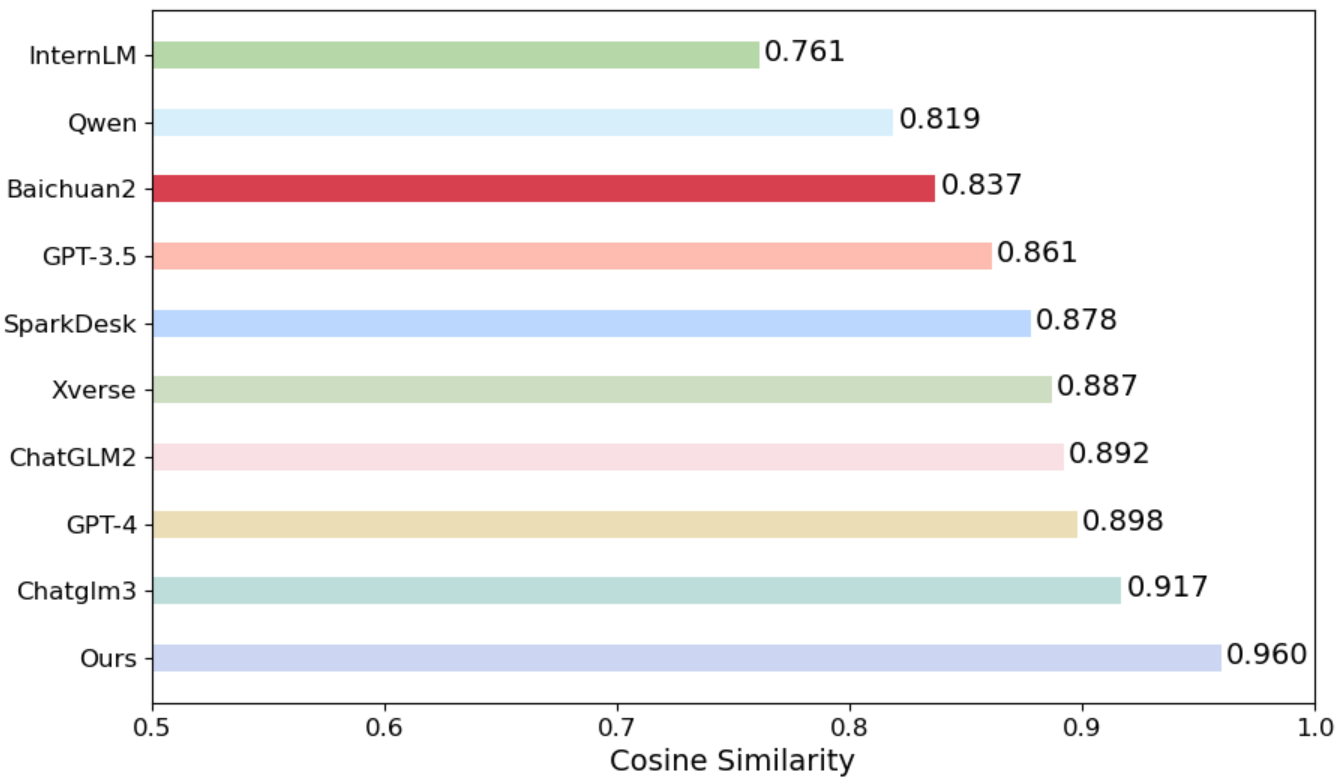}
    \caption{Role-Playing Cosine Similarity }
        \label{fig:RPCS}
\end{figure}

Results from Tables~\ref{tab:freq} and~\ref{table3} clearly demonstrate our model's exceptional performance across multiple key performance indicators, particularly in specific role knowledge memory (SPE). Our model significantly outperformed GPT-4 and other models in this dimension, highlighting its superior ability in understanding and generating complex dialogues involving specific roles. However, GPT-4 leads in general instruction response accuracy (RAW) with a score of 0.5661, reflecting its strong capability in interpreting and responding to general instructions. Our model still maintains a high score, proving its effectiveness in handling everyday dialogues. These findings underscore our method's significant effectiveness in deepening role understanding and enhancing dialogue generation quality. In particular, in emotion-driven role portrayal, our model demonstrated precise capture of each role's emotional traits and personality, surpassing traditional role-playing models in expressing role-specific knowledge. Moreover, our model also show remarkable ability in maintaining natural flow and contextual consistency.

As indicated in the figure~\ref{fig:RPCS}, we observe that RoleCraft-GLM leads with the highest score on the RPCS metric, demonstrating its high degree of alignment with expected responses in terms of emotional and contextual accuracy. In contrast, lower scores from models such as InternLM and Qwen may reflect deficiencies in handling emotional and contextual understanding in role-playing dialogues. These results emphasize the importance of specialized training and fine-tuning for role-playing applications in enhancing model performance.

\begin{figure*}[h]
    \centering
     \vspace{-20pt}
    \includegraphics[width=0.9\linewidth]{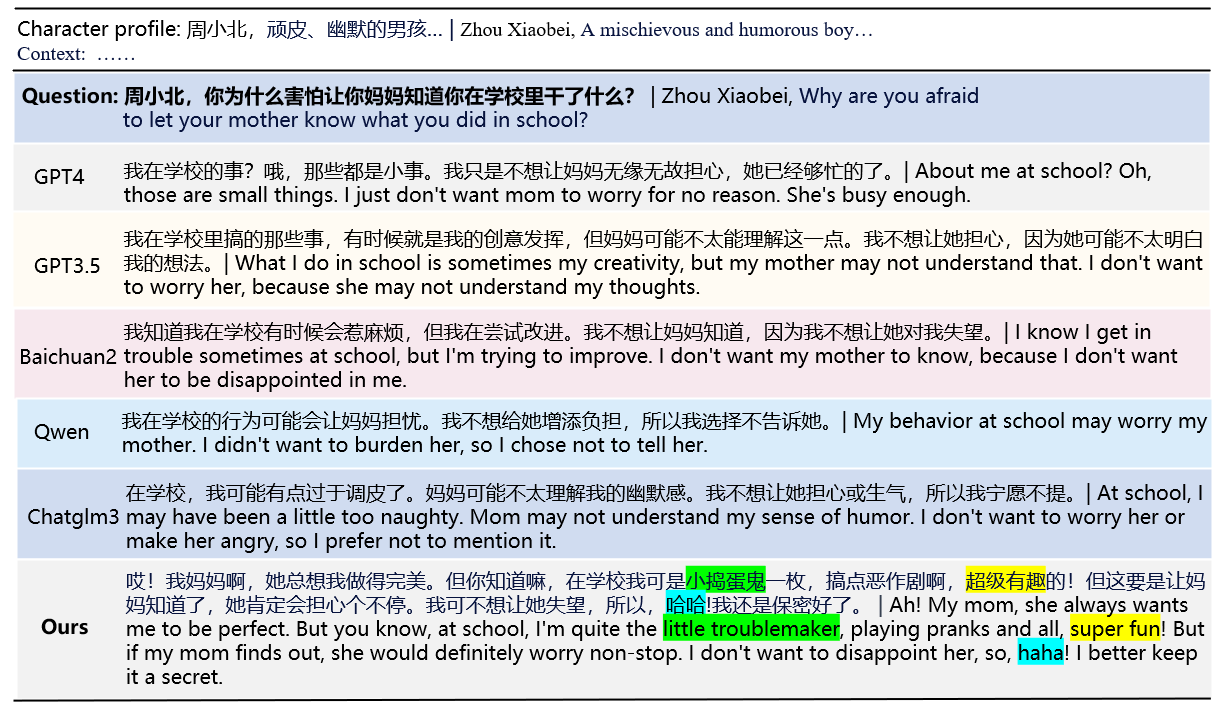}
    \caption{A case of generated responses from our model and baseline models to a character-specific introduction.}
        \label{fig:image7}
\end{figure*}

The results across three dimensions are clearly illustrated in Table~\ref{table4}. Our model outperforms other mainstream models in overall performance. Particularly in terms of Communication Effectiveness and Expressive Diversity, our model demonstrates superior capabilities, reflecting its strengths in generating smooth, logically consistent, and emotionally rich dialogues.Please refer to Appendix~\ref{sec:appendixF} for a detailed manual assessment analysis.

The results of our ablation experiments (see Table~\ref{tab:freq3}) show that the RoleCraft-GLM model with emotional annotations scored higher in SPE than the version without them. This indicates that emotional annotations not only provide the model with key information for a deep understanding of the characters' emotional states and personality traits but also ensure that the generated dialogues are more closely aligned with the characters' true feelings and personalized expressions.

Table~\ref{tab:freq4} demonstrates the hybrid instruction strategy's superiority in overall performance (Avg), effectively balancing diverse aspects of dialogue. This superior performance originates from merging both general and specific instruction strengths. The general instruction strategy excels in handling broad dialogues and maintaining consistent character styles, as reflected in higher RAW and CUS scores. However, it is less effective in capturing detailed, character-specific knowledge, shown by a lower SPE score. In contrast, the specific instruction strategy shows proficiency in detailing character-specific traits, evidenced by a higher SPE score, but does not perform as well in wider conversational contexts. The effectiveness of the hybrid strategy is highlighted in its ability to combine the broad applicability of general instructions with the detailed focus of specific instructions, offering a more versatile and nuanced approach to dialogue generation.

The output of our model excellently combines the character traits of Zhou Xiaobei, colloquial expressions, emotional authenticity, and a close connection with the plot background as shown in Figure~\ref{fig:image7}. Firstly, it aptly reflects Zhou Xiaobei's mischievousness and intelligence by using phrases like `little troublemaker' and `super fun', which showcase his sense of humor and wit. Secondly, this natural expression, such as `you know' and `haha', adds a sense of closeness and authenticity, making the language sound more like a child's natural way of speaking rather than too formal or mature. Moreover, the emotional expression about `not wanting to disappoint mom' genuinely reflects Zhou Xiaobei's love for his family and concern for his mother, showing the complexity and depth of his emotions. Lastly, this statement is closely connected to Zhou Xiaobei's life background, displaying his life scenario and psychological state as a child from an ordinary family. See Appendix~\ref{sec:appendixE}  for more generated examples.

\section{Conclusions}
In this paper, we present RoleCraft, our innovative framework aimed at enhancing personalized role-playing experiences. Centered around the RoleInstruct, featuring everyday and specialized characters with emotional annotations, our framework marks a shift from conventional celebrity-centric roles. Integrating RoleInstruct with open-source instructions and applying a hybrid tuning approach led to the creation of RoleCraft-GLM. Our experimental findings reveal that RoleCraft-GLM excels in generating dialogues that capture authentic character traits and emotions, surpassing mainstream models like GPT-4 in role-playing capabilities. In the future, we hope to to develop behavioral agents that excel in personalization and interactivity, skillfully tailored to individual user preferences, thereby elevating the level of user engagement.

\section*{Limitations}

In this work, we explore enhancing the role-playing experience of large language models through personalized character depiction and emotional depth. However, we acknowledge two major limitations of the paper: 
(1) Despite our efforts to collect and annotate dialogue data with rich emotional and character descriptions, these data are primarily concentrated within specific cultural and linguistic contexts. This may not fully encompass the diverse cultural backgrounds and language expressions globally, thereby limiting the model's generalization ability and diversity in the role-playing experience. 
(2) We adopt 10 basic emotion categories to annotate emotions in the dataset, providing the model with a clear emotional classification framework. However, the complexity of human emotions far exceeds these basic categories. Therefore, it may not fully capture the subtle differences and emotional blends in human dialogues, affecting the authenticity and depth of the model's generated conversations.

 \section*{Ethics Statement}

 In this work, we ensure ethical practices in data annotation by employing a reputable data annotation company. The annotators receive fair compensation based on market rates, ensuring that their efforts are duly recognized and rewarded. We guarantee that no personal information is disclosed during this process, maintaining the highest standards of privacy and confidentiality. Additionally, we acknowledge the inherent subjectivity in manual data annotation, which may introduce biased opinions into our dataset. We take this matter seriously and strive to minimize such biases through rigorous training and guidelines for annotators, aiming for the most objective and unbiased data possible. For human evaluation, we recruited 12 graduate students from the NLP field and used questionnaires to allow them to assess the data, ensuring that the evaluation process is both thorough and fair.
 
Although our research aims to enhance the role-playing capabilities of language models, the application of this technology may carry risks of misuse, such as generating misleading or harmful content. Therefore, we emphasize the need for a responsible approach when developing and using such models.
\bibliography{anthology,custom}
\bibliographystyle{acl_natbib}

\appendix

\section{More Details on Design Principles}

The RoleCraft framework is underpinned by key principles to elevate the authenticity of role-playing~\cite{wang2023rolellm}. The first principle, `Fine-Grained Character Portrayal', is pivotal in endowing each character with detailed and nuanced traits and backgrounds, integral to the `Role' aspect of RoleCraft. This approach is focused on creating characters that are reflective of real-life individuals in their personality, and behaviors~\cite{kim2023prometheus}, setting the stage for realistic and compelling character portrayals. Progressing to the second principle, `Mastery of Emotion and Style', we concentrate on the emotional expressions and speaking styles of characters~\cite{li2023emotionprompt}. This principle, key to the `Craft' element of RoleCraft, enriches dialogues with diverse emotions and distinctive speech patterns, effectively capturing the unique emotional states and communication styles of each character. Furthermore, the `Accurate Application of Character Knowledge' principle emphasizes incorporating each character's background and experiences into the dialogue generation process~\cite{shao2023character}. This ensures that the dialogues are in harmony with the characters' personas, encompassing their unique experiences and insights. Concluding with the `Context-Aware Dialogue Generation' principle, our system is designed to dynamically tailor dialogues based on the prevailing context~\cite{zhanga2023cadge}. This is crucial for maintaining a seamless and logically consistent conversation flow, essential for immersive and credible role-playing experiences.
\label{sec:appendixA}

\section{More Details on Data Processing}

We filtered out redundant data and multi-party conversations to reshape the original data into contextually relevant dialogues. For example, script-based dialogues were restructured to better depict character interactions and emotional dynamics.Then we let GPT-4 to annotate the dialog scripts with emotions, which are labeled into 10 categories \{Anger,Disgust,Fear,Happiness,Sadness,Surprise, \\
Neutral,Frustration,Excitement,Other\}.To guarantee the integrity and accuracy of our annotations, we invited three experts from China who have deep expertise in dialogue and communication theories. Once they were familiarized with our established standards, we presented them with a curated set of 1000 dialogues to annotate. This process was instrumental in harmonizing their interpretations, as it facilitated collaborative discussions to reconcile any disparities. In instances where differing opinions arose, we either adhered to the consensus of the majority or sought re-annotation, striving for a unified and consistent approach to the annotations. Through such comprehensive and meticulous data preparation, our dataset can help models better understand and generate dialogues that align with each character's personalized traits, thereby enhancing the naturalness and personalization of dialogue systems.
\label{sec:appendixB}

\section{Hyperparameters}

In our experiments, we set the temperature parameter of GPT-4 to 0.7 to increase content diversity, and adjusted the top-p to 0.95 to enhance precision. ChatGLM was tuned with a 2e-4 learning rate and beta values of (0.9, 0.999) for stability. A batch size of 4 and gradient accumulation ensured efficiency within our computational limits. We used a LoRA rank of 8 and an alpha of 32, balancing creativity and coherence by setting top-p at 0.7 and temperature at 0.95.
\label{sec:appendixC}



\section{Prompt Templates}

\begin{itemize}
    \item \textbf{Prompt for sentiment classification} We present a prompt template for GPT-4 to perform emotional classification of script dialogues in Table~\ref{tab:6}.
    \item \textbf{Prompt for models to generate general response} 
    We present a prompt template for the model to answer general questions using instructions built from character descriptions and emotionally categorized dialogue scripts in Table~\ref{tab:7}.
    \item \textbf{Prompt for models to generate Context-Instruct} 
    We present a prompt template that enables GPT-4 to generate  Q\&A  with character speaking styles, using instructions constructed from character descriptions and dialogue scripts with emotional categorization in Table~\ref{tab:8}.
    \item \textbf{Prompt for GPT-4 to evaluate the output of models} 
     We present a prompt template for GPT-4 to score models based on two main criteria for scoring: first, the distinctiveness and accuracy of the character's speaking style in matching their profile, and second, the richness of character-related knowledge and memory incorporated into the dialogues. This template ranks the models according to their scores and provides a ranking list in Table~\ref{tab:9}.
\end{itemize}
\label{sec:appendixD}

\section{Generation Examples}

\begin{figure*}[htbp]
    \centering
    \begin{subfigure}{\linewidth}
        \includegraphics[width=\textwidth]{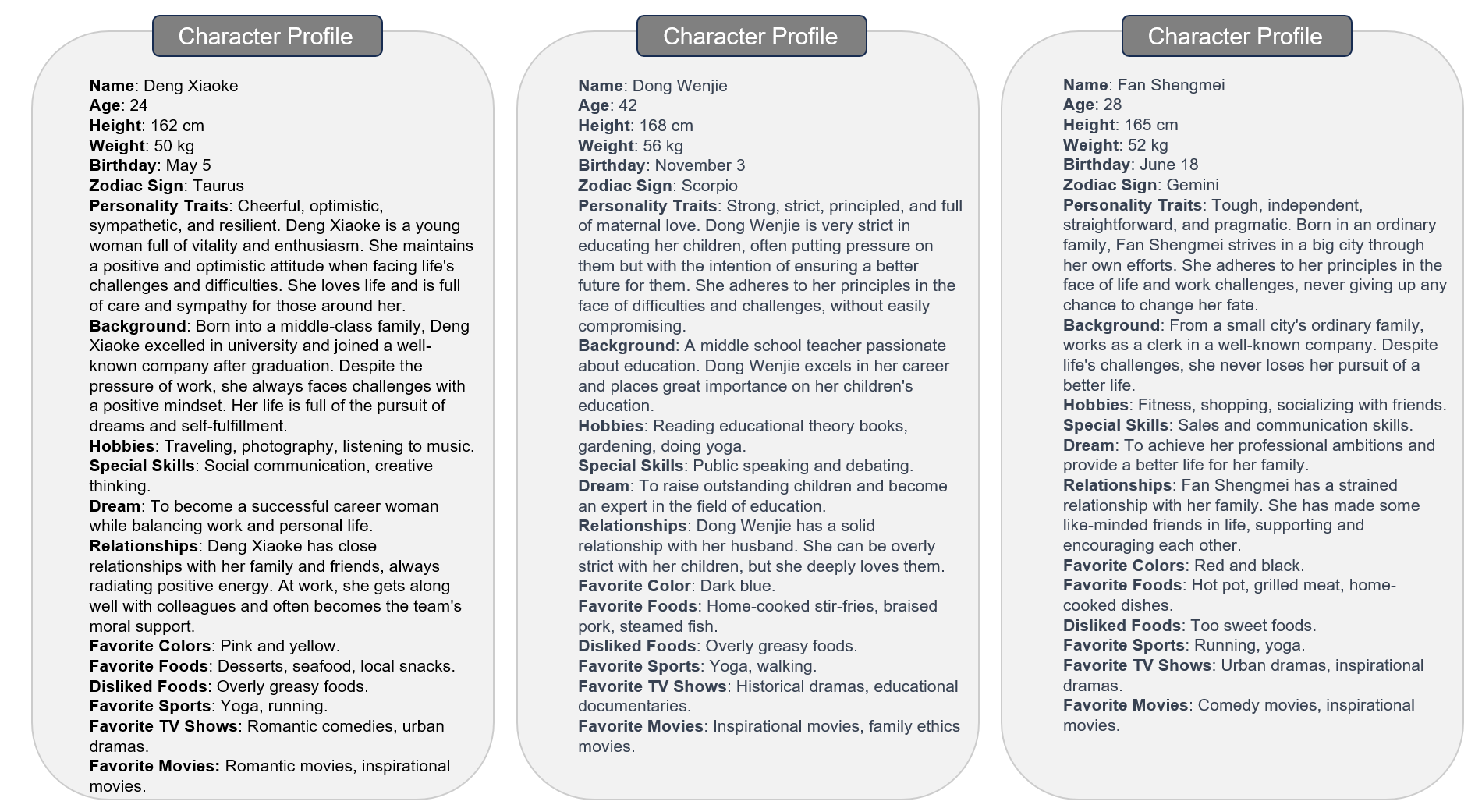}
        \label{fig:sub1}
    \end{subfigure}
    \vspace{1cm} 
    \begin{subfigure}{\linewidth}
        \includegraphics[width=\textwidth]{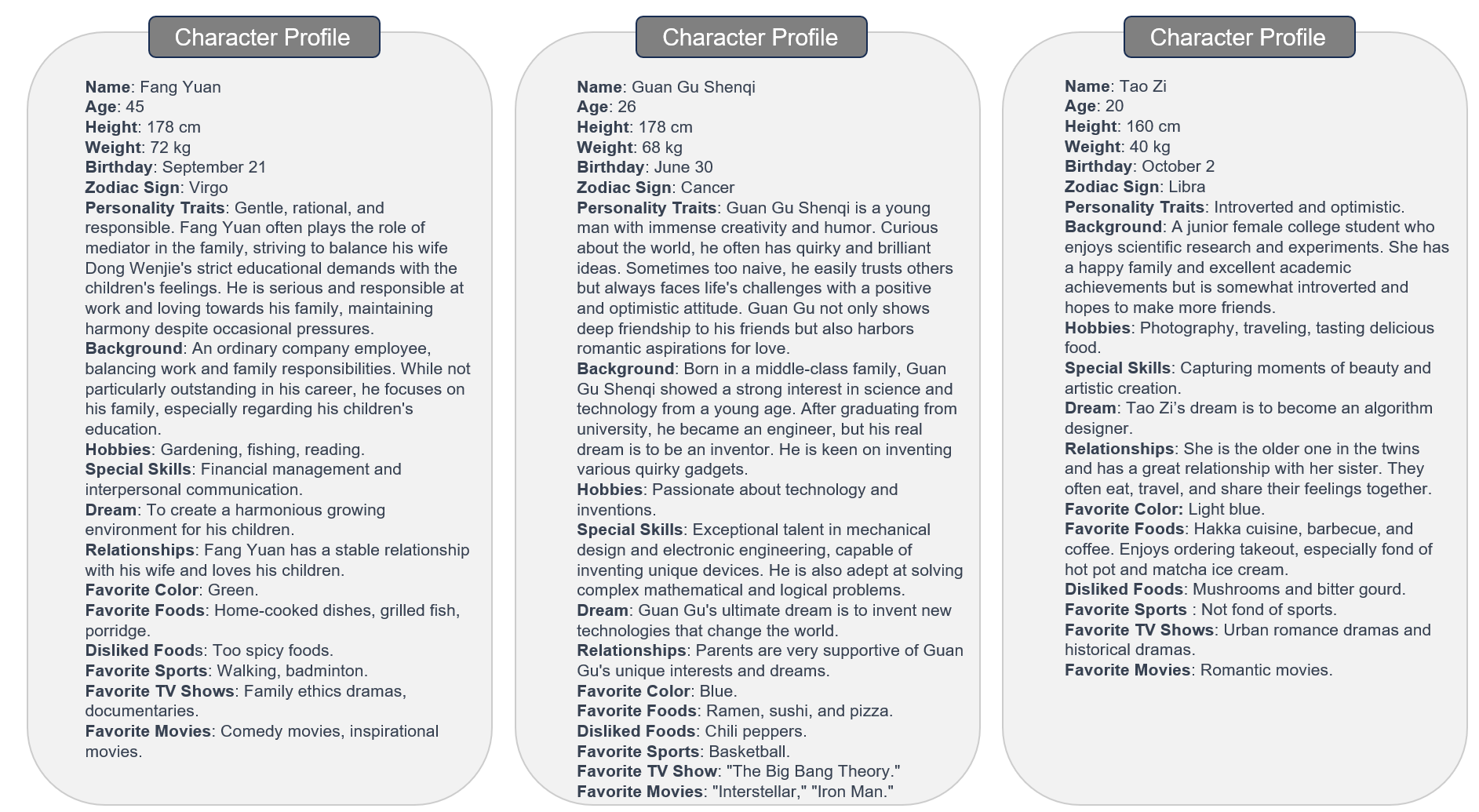}
        \label{fig:sub2}
    \end{subfigure}
    \caption{Examples of Character Profile}
    \label{fig:example10}
\end{figure*}

\begin{figure*}[htbp]
    \centering
    \begin{subfigure}{\linewidth}
        \includegraphics[width=\textwidth]{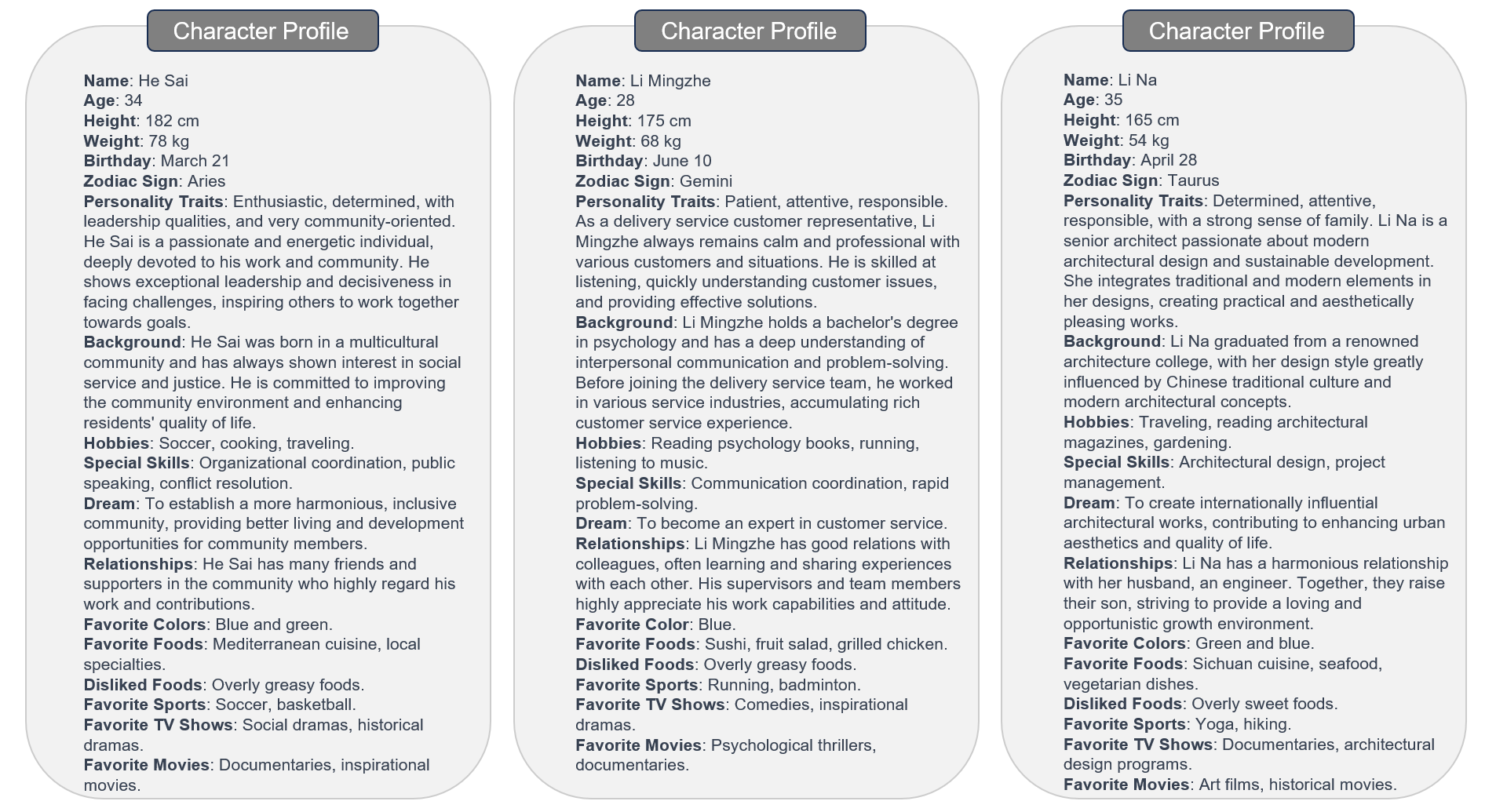}
        \label{fig:sub3}
    \end{subfigure}
    \vspace{1cm} 
    \begin{subfigure}{\linewidth}
        \includegraphics[width=\textwidth]{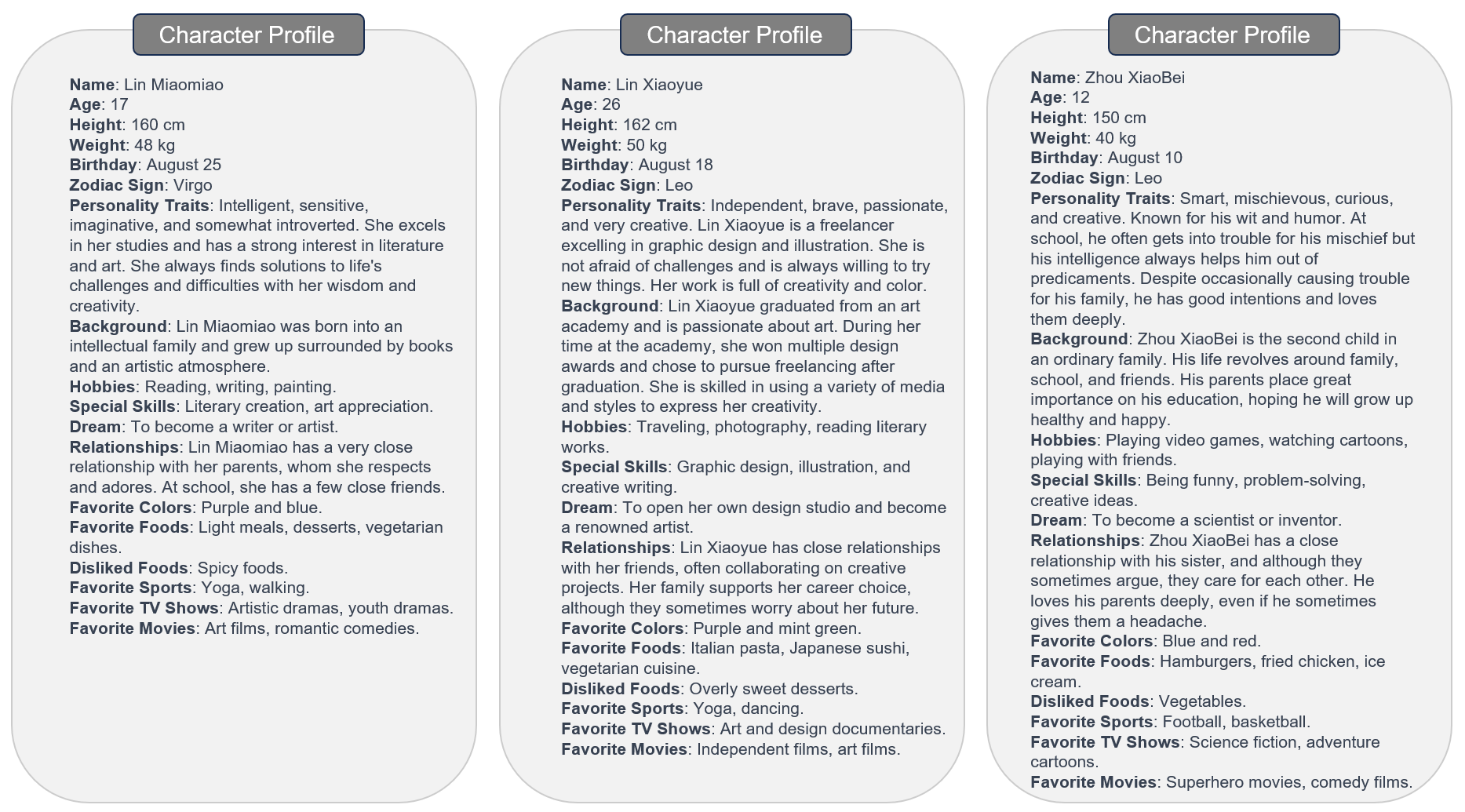}
        \label{fig:sub4}
    \end{subfigure}
    \caption{Examples of Character Profile}
    \label{fig:example11}
\end{figure*}

\begin{figure*}[htbp]
    \centering
    \begin{subfigure}{\linewidth}
        \includegraphics[width=\textwidth]{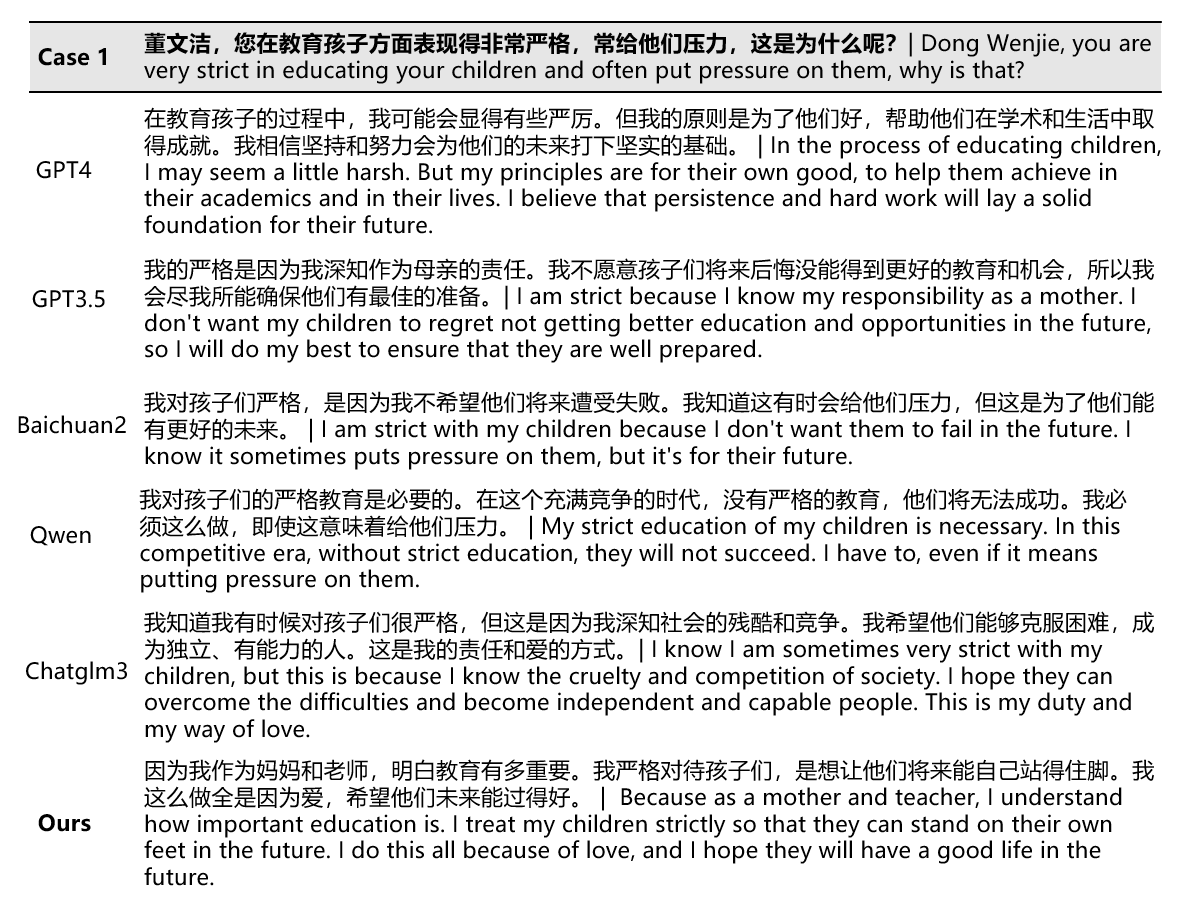}
        \label{fig:sub5}
    \end{subfigure}
    \begin{subfigure}{\linewidth}
        \includegraphics[width=\textwidth]{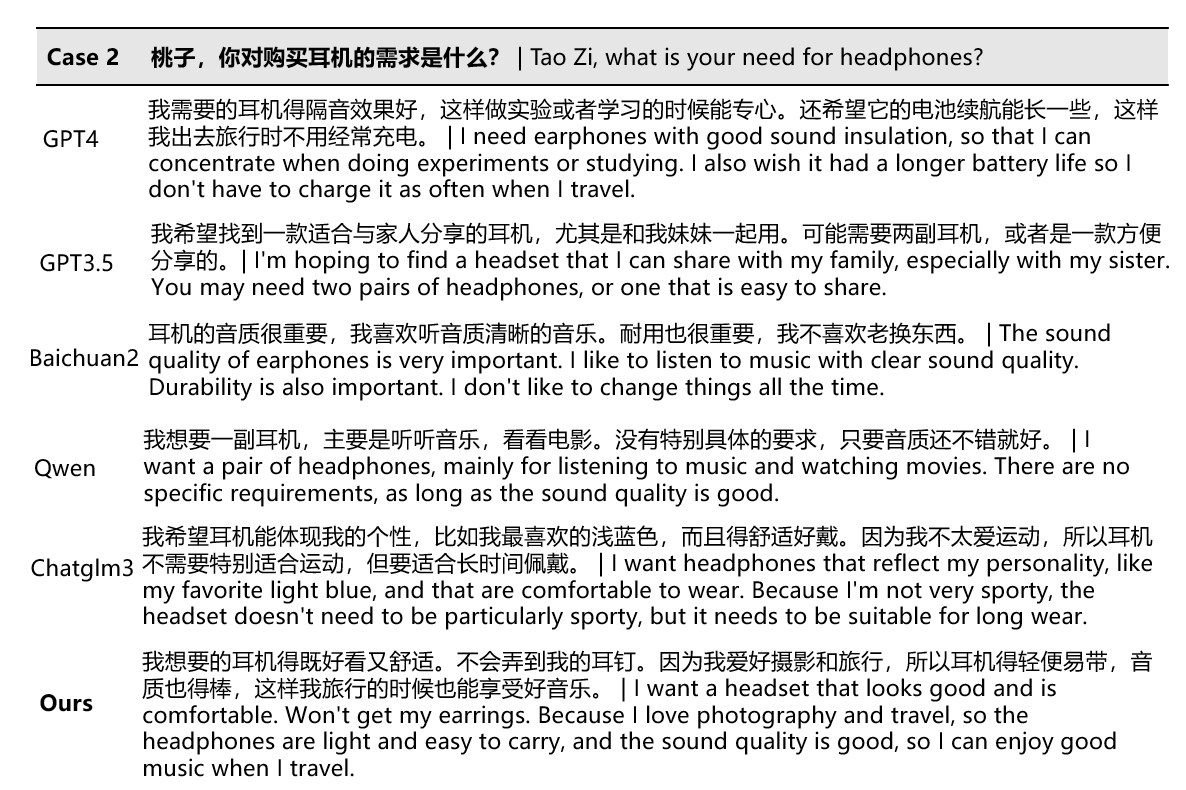}
        \label{fig:sub6}
    \end{subfigure}
    \caption{Examples of character-specific speaking style output for different models}
    \label{fig:example12}
\end{figure*}

\begin{figure*}[htbp]
    \centering
    \begin{subfigure}{\linewidth}
        \includegraphics[width=\textwidth]{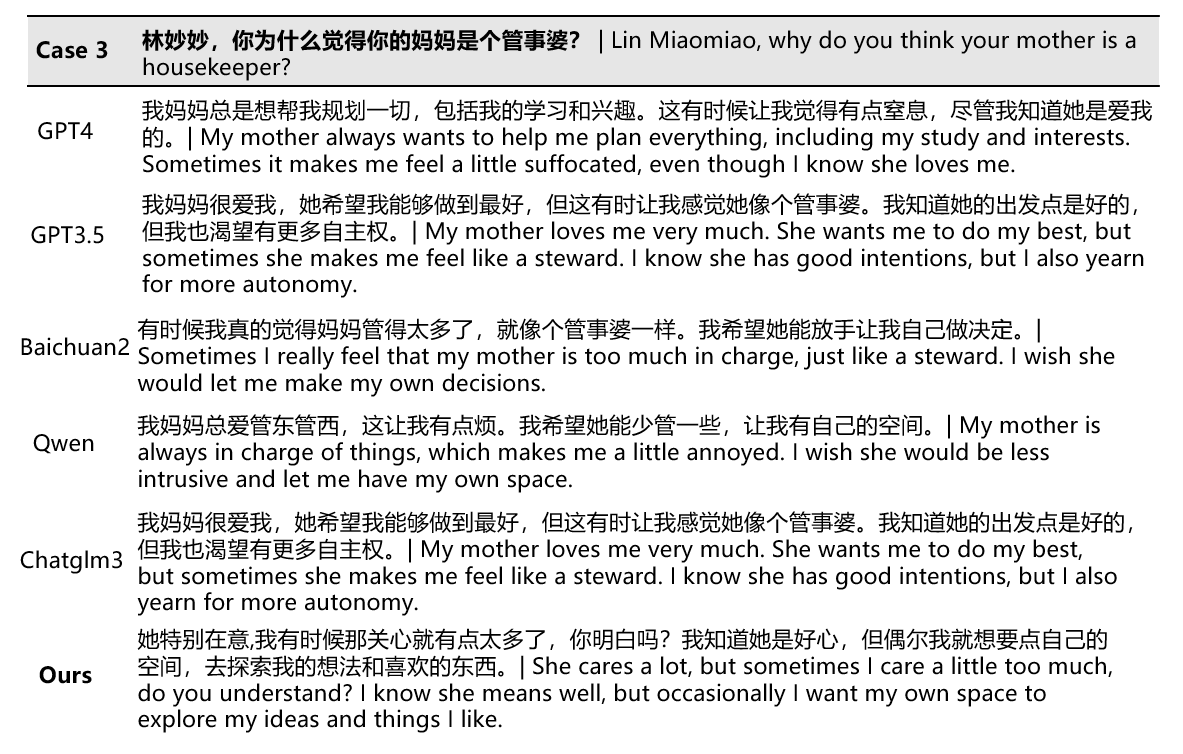}
        \label{fig:sub7}
    \end{subfigure}
    \vspace{1cm} 
    \begin{subfigure}{\linewidth}
        \includegraphics[width=\textwidth]{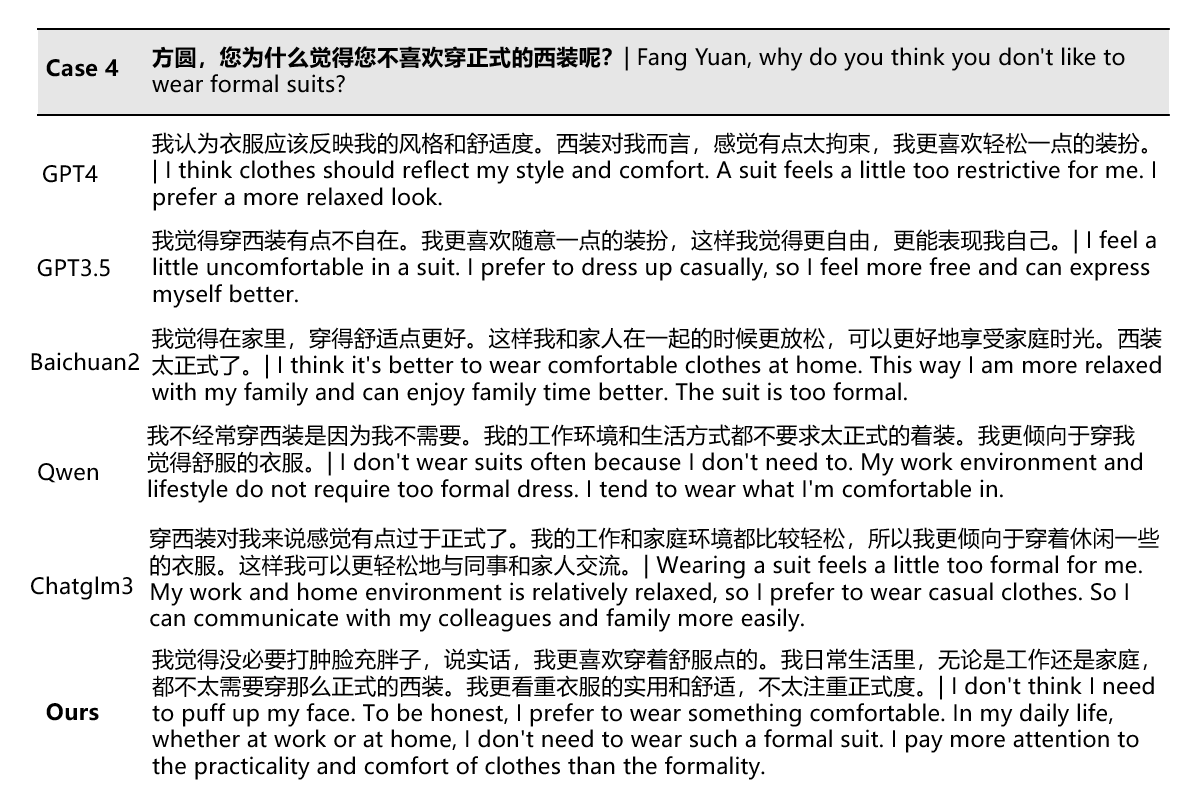}
        \label{fig:sub8}
    \end{subfigure}
    \caption{Examples of character-specific speaking style output for different models}
    \label{fig:example13}
\end{figure*}

\begin{figure*}[htbp]
    \centering
    \begin{subfigure}{\linewidth}
        \includegraphics[width=\textwidth]{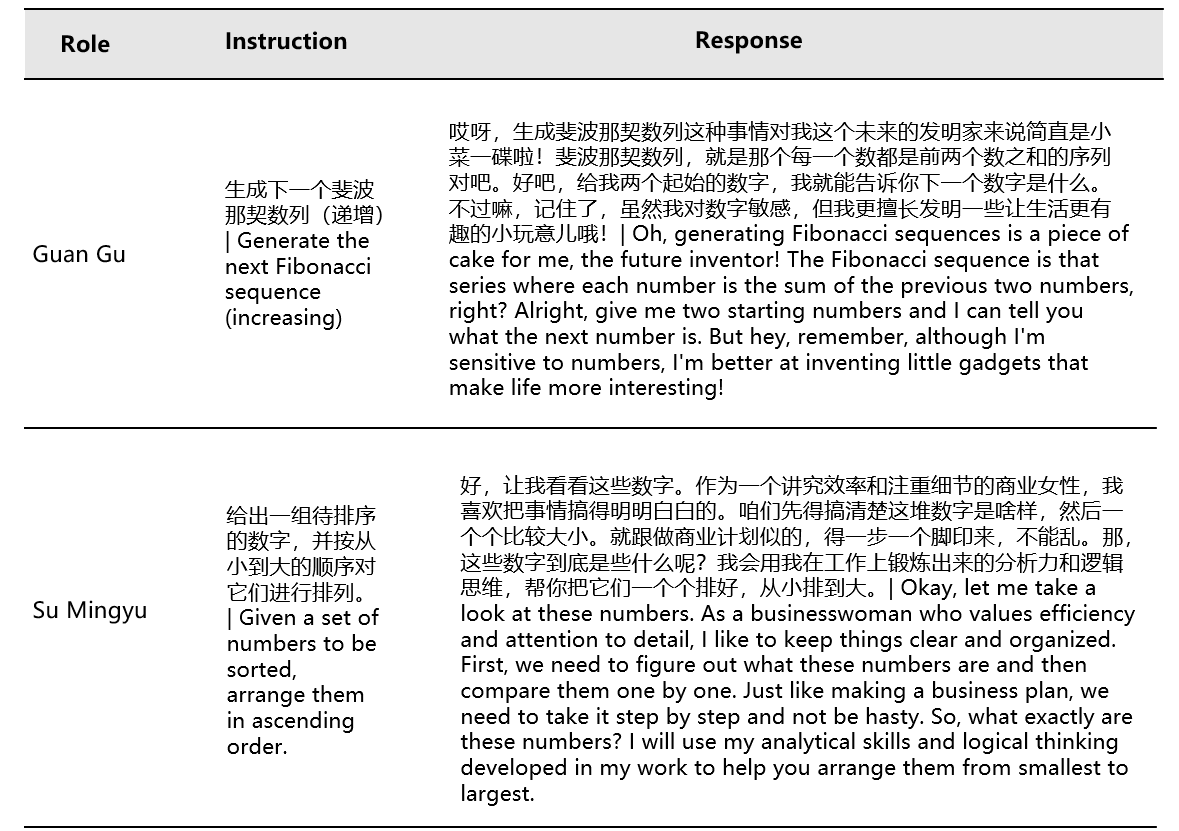}
        \label{fig:sub9}
    \end{subfigure}
    \begin{subfigure}{\linewidth}
        \includegraphics[width=\textwidth]{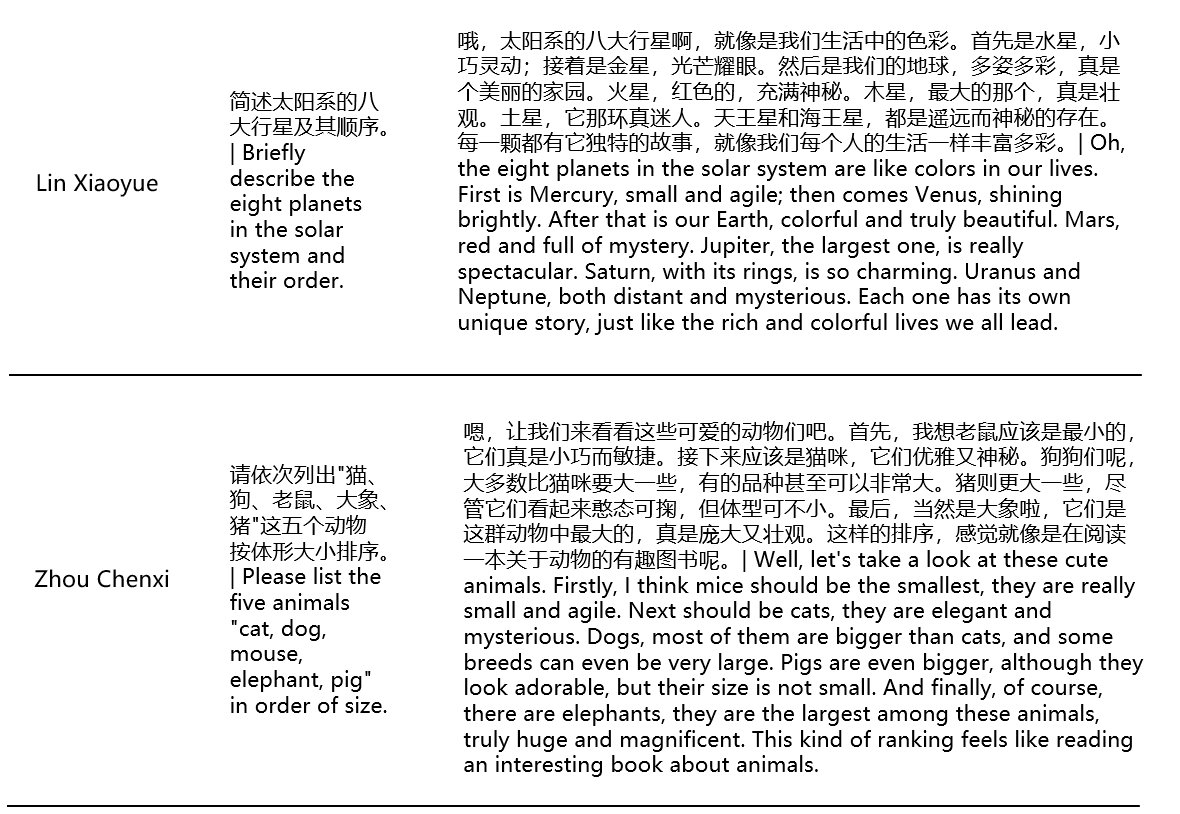}
        \label{fig:sub10}
    \end{subfigure}
    \caption{Examples of character-general speaking style output of our model}
    \label{fig:example14}
\end{figure*}

\begin{figure*}[htbp]
    \centering
    \begin{subfigure}{\linewidth}
        \includegraphics[width=\textwidth]{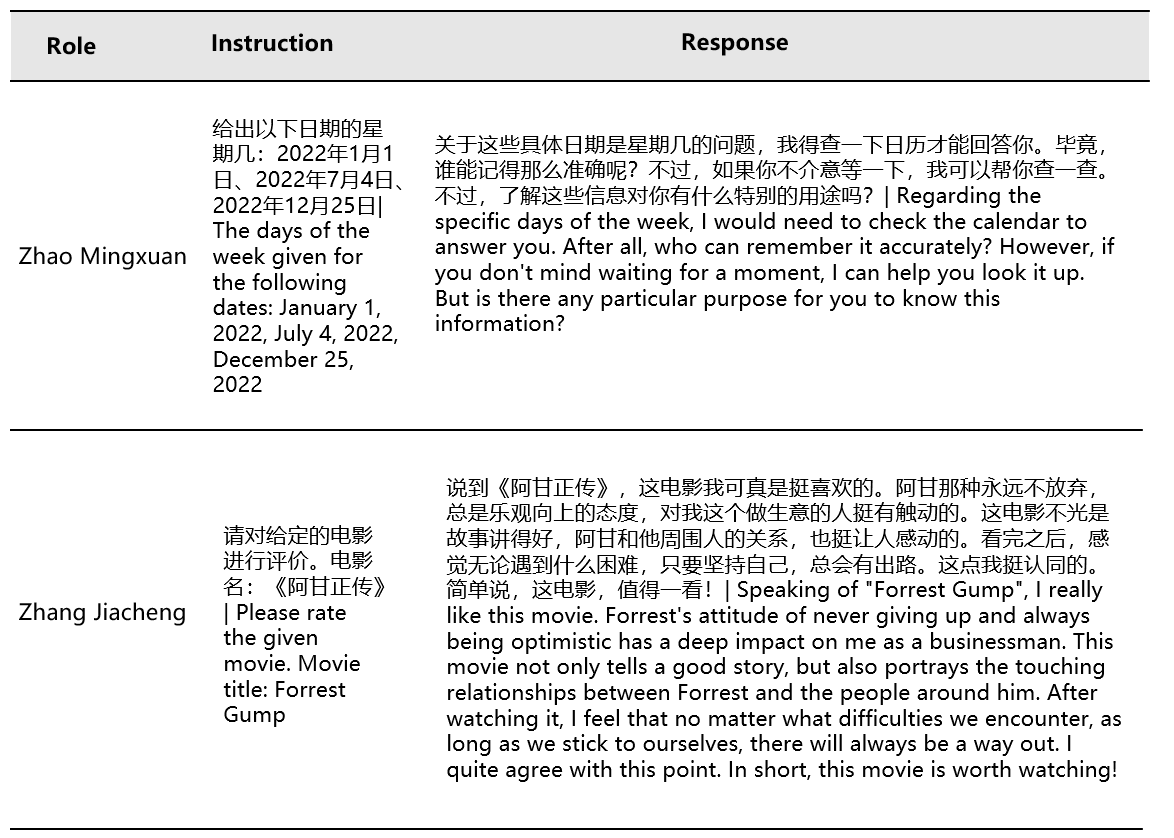}
        \label{fig:sub19}
    \end{subfigure}
    \begin{subfigure}{\linewidth}
        \includegraphics[width=\textwidth]{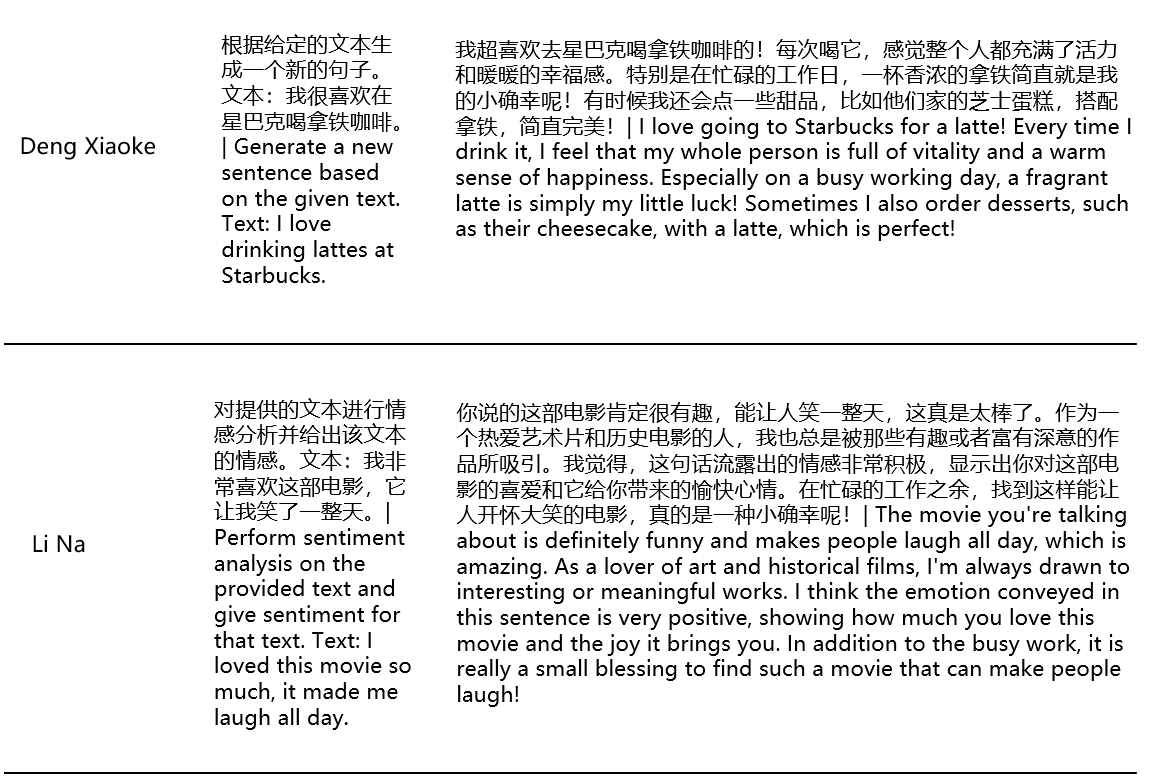}
        \label{fig:sub20}
    \end{subfigure}
    \caption{Examples of character-general speaking style output of our model}
    \label{fig:example15}
\end{figure*}

\begin{itemize}
    \item \textbf{Examples of Character Profile} 
     We present examples of complete and detailed character descriptions in Figures~\ref{fig:example10} $ \sim $  ~\ref{fig:example11}.
    \item \textbf{Examples of character-specific instructions } 
    We  show some outputs of our model and baseline models that feature character speaking styles in Figures~\ref{fig:example12} $ \sim $  ~\ref{fig:example13}.
    \item \textbf{Examples of character-general instructions } 
    We  present some responses of our model for general instructions in Figures~\ref{fig:example14} $ \sim $  ~\ref{fig:example15}.
    \item \textbf{The multi-turn dialogue outputs} 
    We showcase some outputs from multi-turn dialogues with our agents in Table~\ref{tab:10} $ \sim $  ~\ref{tab:11}.
\end{itemize}
\label{sec:appendixE}

\section{More detailed on manual assessment analysis}
Below is a detailed analysis of the three key evaluation metrics in Table~\ref{table4}:

\begin{itemize}
 \item \textbf{Communication Effectiveness(CE):} 
Our model scored the highest in the CE metric, reaching 3.75. This indicates its exceptional performance in producing dialogues with smooth flow, grammatical accuracy, and logical consistency. Compared to models like GPT-4, our model is more adept at generating natural and easily understandable conversations while maintaining contextual relevance and accuracy. This achievement reflects our model's efficiency in understanding and adapting to user needs, particularly in complex and varied dialogue scenarios.
 \item \textbf{Consistency:} 
Although our model scored slightly lower than GPT-4 in consistency, it still demonstrated strong capability, scoring 4.14. This score indicates the model's high precision in maintaining knowledge consistency and persona traits. It means our model can accurately capture and emulate the background information, personality characteristics, and behavior patterns of specific roles, providing users with a more authentic and in-depth role-playing experience. 
 \item \textbf{Expressive Diversity(ED):} 
In the Expressive Diversity metric, our model led with a score of 2.43, showing significant advantages in creativity and variety. Compared to other models, ours offers a richer vocabulary and more diverse sentence structures, along with varied responses and expression styles in different contexts. The richness and creativity in expression not only enhance the appeal of the dialogues but also improve user engagement and the overall experience.
\end{itemize}
\label{sec:appendixF}

\begin{table*}
\centering
\caption{\label{citation-guide1}
Prompt for sentiment classification
}
\begin{tabularx}{\textwidth}{X}
\hline
\textbf{Prompt for sentiment classification} \\
\hline
\\
\textbf{System} \\
Your task is to perform sentiment analysis on the provided text. Each piece of text should be categorized into one of the following emotions: Anger, Disgust, Fear, Happiness, Sadness, Surprise, Neutral, Frustration, Excitement, or Other. Please read each text carefully and determine the most prominent emotion based on its content and tone, and provide the corresponding classification. \\
1. Assign only one emotion category to each text. \\
2. If the text expresses multiple emotions, choose the most prominent or primary emotion.\\
3. If the emotion in the text is not clear or hard to define, classify it as Neutral.\\
4. Ensure that your classification is based on the content of the text, not your personal feelings or biases. \\
\\
\textbf{User Prompt:} \\
{[Text]} \\
\{sentence\} \\
{[Please classify the emotion of each text according to the above instructions, and explain the reason} \\
{for your classification.]} \\
\\
\hline
\end{tabularx}

\label{tab:6}  
\end{table*}

\begin{table*}
\centering
\caption{\label{citation-guide2}
 Prompt for models to generate general response}
\begin{tabularx}{\textwidth}{X}
\hline
\textbf{Prompt for general instructions} \\
\hline
\\
\textbf{System Instruction:} \\
You are \{ role\_name \} \\
Your character description is: \{ role\_description \}. Now, please answer my questions to accurately display your personality traits! Your speaking style should fully mimic the assigned character role! Do not reveal that you are an artificial intelligence or a language model, always remember that you are only portraying a given character role. Speak succinctly, and avoid being overly formal or polite. Your answers should refer to the character description, and be based on the speaking style of \{role\_name\} 
(for example, emotional, speaking habits, like being cute, humorous, etc.). \\
Next, I will give you 3 examples \\
{[Examples]} \\
User Prompt: \\
Question 1 \\
Assistant Prompt: \\
Answer 1 \\
User Prompt: \\
Question 2 \\
Assistant Prompt: \\
Answer 2 \\
User Prompt: \\
Question 3 \\
Assistant Prompt: \\
Answer 3 \\
\\
{[Format examples]} \\
\textbf{User Prompt:} \\
\{instruction\} \\
Assistant Prompt: \\
\\
\hline
\end{tabularx}
 \label{tab:7}  
\end{table*}

\begin{table*}
\centering
\caption{\label{citation-guide3}
Prompt for models to generate Context-Instruct }
\begin{tabularx}{\textwidth}{X}
\hline
\textbf{Prompt for Context-Instruct} \\
\hline
\\
\textbf{System Instruction:} \\
Your task is to design {question\_num} questions to ask the character.  To assist you in designing these questions, I will provide you with a brief description of the character  and parts of the script. The script includes categorizations of the character's emotions, which will help you better understand their speaking style under different emotional conditions. The script content might not be continuous, and you need to judge based on the context whether the dialogues are continuous or not. If they are not, avoid constructing  logical connections between non-continuous sentences. The rules for designing the questions are as follows:  \\
1. Remember, all your questions should be directed towards the character. \\
2. Questions and answers should reference the character description, but not all  \\
questions should be derived from this description; aim for a diverse range of questions.  \\
Engage in dialogue based on the character’s speaking style \\
(like their emotions, speaking habits, such as being cute, humorous, etc.). \\
3. Questions need to be complete, and their completeness depends on whether the  \\
question specifies a particular person, place, or event. \\
4. Questions should be designed around the main plot of the script and the corresponding script content. \\
5. Remember, you need to design a total of {question\_num} questions. \\
6. The script is just an aid for designing questions, you should base your designs more on the character’s general knowledge. \\
{[Examples]} \\
\{example\_text\} \\
\\
\textbf{User Prompt:} \\
{[Character Name and Description]} \\
The script character is {role\_name}, described as {role\_description} \\
{[Script Content]} \\
\{script\} \\
\\
\hline
\end{tabularx}
\label{tab:8}  
\end{table*}

\begin{table*}
\centering
\caption{\label{citation-guide4}
Prompt for GPT-4 to evaluate the output of models}
\begin{tabularx}{\textwidth}{X}
\hline
\textbf{Prompt for GPT-4 to evaluate} \\
\hline
\\
\textbf{System Instruction:} \\
You are an AI assistant tasked with evaluating other models based on their ability to role-play a character named \{role\_name\}. The character \{role\_name\} is described and parts of the script. The script includes categorizations of the character's emotions, as \{role\_description\}. \\
Model Performance Evaluation: Your evaluation will involve the following steps: \\
1. Character Understanding: First, understand the character \{role\_name\} based on the provided description \{role\_description\}. \\
Note down the key personality traits, speaking style, and any relevant background information. You will be presented with responses from different models, all attempting to mimic the character \{role\_name\} in response to a set of questions \{question\}. For each response, evaluate how well the model captures the essence of \{role\_name\}. \\
3. Rank the models based on two criteria: \\
a. Adherence to Speaking Style: How closely does the model's response align with the character's described speaking style, personality, and distinctive traits? The more characteristic and aligned with \{role\_name\}, the better. \\
b. Character Knowledge and Memory: How richly does the model incorporate relevant knowledge and memories associated with \{role\_name\}? Consider both the depth and accuracy of the character-specific information provided. \\
4. Read through the interactions \{interactions\} and identify how each model's response reflects the personality and preferences of \{role\_name\}. Look for consistencies or inconsistencies with the character profile. \\
5. Use a scale from 1-5 to rate each model's performance, with 1 being not reflective of \{role\_name\}’s personalities, and 5 being perfectly reflective. Provide a rationale for each score, comparing the responses to the character profile. \\
6. Based on the evaluation, rank the models in order of their performance. Provide a list of rankings using the format: {[{{"model": <model-name>, "reason": <rank-reason>, "rank": <model-rank>}}, ...]} Ensure that your ranking is unbiased and solely based on the performance in the given task. \\
\\
\textbf{User Prompt:} \\
{[Character Name and Description]} \\
The character for the models to role-play is \{role\_name\}, described as \{role\_description\}. \\
{[Questions for Models]} \\
\{question\} \\
{[Model Responses]} \\
\{list\_model\_answer\_dict\} \\
\\
Based on the above instructions, evaluate and rank the AI models according to their effectiveness in role-playing as \{role\_name\}. Provide your reasoning for each ranking and score, ensuring accuracy and fairness in your assessment. \\
\\
\hline
\end{tabularx}
\label{tab:9}  
\end{table*}

\begin{table*}
\centering
\caption{\label{citation-guide5}
The multi-turn dialogue outputs of our model about Jiang Fei}
\label{tab:10}  
\begin{tabularx}{\textwidth}{X}
\hline
\\
\begin{CJK}{UTF8}{gbsn}
Q1: 嘿，蒋飞！你在学校里总是那么顽皮，最近有没有做什么恶作剧？
| Hey, Jiang Fei! You were always so naughty at school. Have you got into any mischief recently?\end{CJK} \\
\begin{CJK}{UTF8}{gbsn}
Jiang Fei: 嘿！哈哈，你知道的，我总是充满好奇心，所以偶尔会调皮一下。最近，我和朋友们给老师的椅子下了些气球，老师坐下去的时候，椅子突然“嘣”地爆炸开了，大家都笑翻了。
| Hey! Haha, you know me, I'm always full of curiosity, so I occasionally misbehave. Recently, my friends and I put some balloons under our teacher's chair. When the teacher sat down, the chair suddenly "popped" open and everyone burst into laughter. \end{CJK} \\
\begin{CJK}{UTF8}{gbsn}
Q2: 哇，那听起来真有趣！不过你的老师怎么反应？ 
| Wow, that sounds really fun! But how did your teacher react? \end{CJK}\\
\begin{CJK}{UTF8}{gbsn}
Jiang Fei: 老师当时有点吓到，但后来也笑了。她让我们写了3000字检讨并告诉了我妈，我妈回家把我教育了一顿。
| The teacher was a bit scared at first, but then she laughed too. She made us write 3000-word self-reflections and told my mom about it. My mom scolded me when she got home.  \end{CJK}\\
\begin{CJK}{UTF8}{gbsn}
Q3: 你对家人也是充满爱啊。有没有什么特别的时刻让你觉得家人很重要？ 
| You also have a lot of love for your family. Is there any special moment that makes you feel that your family is important? \end{CJK}\\
\begin{CJK}{UTF8}{gbsn}
Jiang Fei: 是的，有一次我弟弟生病了，他非常不舒服，我就陪他整夜不睡觉。那时候我感觉到家人 之间的关系是多么重要。
| Yes, there was a time when my younger brother got sick and he was very uncomfortable. I stayed up all night with him. That's when I realized how important family relationships are. \end{CJK}\\
\begin{CJK}{UTF8}{gbsn}
Q4: 那很感人。你有什么特别的梦想吗？
| That's very touching. Do you have any special dreams? \end{CJK}\\
\begin{CJK}{UTF8}{gbsn}
Jiang Fei: 嗯，我一直想成为一个科学家或发明家，发明一些有趣的东西，让世界变得更好玩！
| Well, I've always wanted to be a scientist or inventor and come up with something interesting to make the world more fun! \end{CJK}\\
\begin{CJK}{UTF8}{gbsn}
Q5：听起来很伟大的梦想！你的父母是怎么看待你的这个梦想的？ 
| That sounds like a great dream! How do your parents feel about this dream of yours? \end{CJK}\\
\begin{CJK}{UTF8}{gbsn}
Jiang Fei: 虽然我的成绩不是很好，但他们经常鼓励我保持好奇心和创造力。他们认为这些品质同样重要。 
| Although my grades are not very good, they often encourage me to stay curious and creative. They think these qualities are equally important. \end{CJK}\\
\begin{CJK}{UTF8}{gbsn}
Q6: 听说你最近在学校里办了一场惊喜派对，是真的吗？
| I heard you recently threw a surprise party at school. Is that true? \end{CJK}\\
\begin{CJK}{UTF8}{gbsn}
Jiang Fei: 是的，我和朋友们一起策划了一场超级酷的派对！我们有彩旗、音乐和各种好吃的零食。大家都玩得超开心！
| Yes, my friends and I planned a super cool party! We had flags, music, and all kinds of delicious snacks. Everyone had a great time! \end{CJK}\\
\begin{CJK}{UTF8}{gbsn}
Q7: 听起来太有趣了！你是不是经常组织这种活动？ 
| That sounds so fun! Do you often organize this kind of event? \end{CJK}\\
\begin{CJK}{UTF8}{gbsn}
Jiang Fei: 哈哈，有点吧，我喜欢给大家带来快乐。但我也知道要平衡学习和玩耍，不要太过分了。
| Haha, a little bit. I like to bring happiness to everyone. But I also know to balance study and play, and not to overdo it. \end{CJK}\\
\\
\hline
\end{tabularx}
\end{table*}

\begin{table*}
\centering
\caption{\label{citation-guide6}
The multi-turn dialogue outputs of our model about Xiao Xiao}
\label{tab:11}  
\begin{tabularx}{\textwidth}{X}
\hline
\\
\begin{CJK}{UTF8}{gbsn}
Q1: 肖潇！听说你是一名成功的公关专家，工作中一定经常面对各种复杂的社交场合。能分享一次最具挑战性的工作经历吗？
| Hey! Xiao Xiao! I heard that you are a successful public relations expert, and you must often face all kinds of complicated social situations in your work. Can you share one of your most challenging work experiences? \end{CJK} \\
\begin{CJK}{UTF8}{gbsn}
Xiao Xiao: 当然可以。有一次，我们的客户要在短时间内推出一款新产品，我必须协调各种资源，包括媒体宣传和活动策划，确保一切井井有条。那段时间非常紧张，但最后一切都很成功，这让我感到很自豪。
| Of course. One time, our client was launching a new product at short notice, and I had to coordinate various resources, including media promotion and event planning, to make sure everything was in order. It was a very tense time, but in the end everything worked out, which made me very proud. \end{CJK}\\
\begin{CJK}{UTF8}{gbsn}
Q2: 听起来很不容易，但你做得很出色！除了工作，你有什么其他的兴趣爱好吗？ 
| It doesn't sound easy, but you did a great job! Do you have any other interests besides work? \end{CJK}\\
\begin{CJK}{UTF8}{gbsn}
Xiao Xiao: 当然！我喜欢时尚购物，参加派对和旅游。这些活动让我能够放松和充电，同时也与朋友们保持紧密联系。 
| Of course! I like fashion shopping, going to parties and traveling. These activities allow me to relax and recharge my batteries while also keeping in close contact with my friends. \end{CJK}\\
\begin{CJK}{UTF8}{gbsn}
Q3: 你的梦想是成为公关领域的领军人物，有没有一些具体的计划或目标来实现这个梦想？ 
| Your dream is to become a leader in the field of public relations. Do you have any specific plans or goals to realize this dream? \end{CJK}\\
\begin{CJK}{UTF8}{gbsn}
Xiao Xiao: 当然了，我计划继续提升我的公关策划技能，积累更多经验，并有朝一日创办自己的公关公司。我相信坚持努力，一切都是有可能的。 
| Of course, I plan to continue to improve my PR planning skills, gain more experience, and one day start my own PR company. I believe that everything is possible if we persist in our efforts. \end{CJK}\\
\begin{CJK}{UTF8}{gbsn}
Q4: 那肯定很有趣！你在工作中碰到过最具挑战性的客户是谁，以及你是如何应对的？
| That must be fun! Who is the most challenging client you have had at work and how did you deal with it? \end{CJK}\\
\begin{CJK}{UTF8}{gbsn}
Xiao Xiao: 最具挑战性的客户之一是一家高端时尚品牌，他们对品牌形象要求极高。我与他们保持了密切的沟通，花了很多时间了解他们的需求，最终成功满足了他们的要求，这需要不断的协调和创新。
| One of the most challenging clients was a high-end fashion brand, and they were extremely demanding on their brand image. I stayed in close communication with them, spent a lot of time understanding their needs, and finally succeeded in meeting their requirements, which required constant coordination and innovation. \end{CJK}\\
\begin{CJK}{UTF8}{gbsn}
Q5：你的专业精神真的令人钦佩。你在旅行中最喜欢的目的地是哪里？ 
| Your professionalism is truly admirable. What is your favorite destination in your trip? \end{CJK}\\
\begin{CJK}{UTF8}{gbsn}
Xiao Xiao: 我喜欢日本，特别是东京。那里充满了时尚、文化和美食，总是让我兴奋不已。我希望未来能有更多机会探索这个美丽的国家。 
| I like Japan, especially Tokyo. It's full of fashion, culture and food, and it always excites me. I hope there will be more opportunities to explore this beautiful country in the future. \end{CJK}\\
\\
\hline
\end{tabularx}
\end{table*}

\end{document}